\documentclass[a4paper,fleqn]{cas-sc}

\usepackage[numbers, square]{natbib}

\usepackage{amsmath}
\usepackage[linesnumbered,boxed]{algorithm2e}
\usepackage{caption}
\usepackage{subcaption}
\usepackage{graphicx}
\usepackage{longtable}
\usepackage{multirow}
\usepackage{array}
\usepackage{pdflscape}
\usepackage{booktabs}
\usepackage[figuresright]{rotating}
\usepackage{multirow}

\newproof{pf}{Proof}
\usepackage{lineno}
\usepackage{float}
\def\tsc#1{\csdef{#1}{\textsc{\lowercase{#1}}\xspace}}
\tsc{WGM}
\tsc{QE}
\tsc{EP}
\tsc{PMS}
\tsc{BEC}
\tsc{DE}

\begin{document}
	% \linenumbers
	\let\printorcid\relax
	\let\WriteBookmarks\relax
	\def\floatpagepagefraction{1}
	\def\textpagefraction{.001}

	\title [mode = title]{A Joint Topology-Data Fusion Graph Network for Robust Traffic Speed Prediction with Data Anomalism}

	\author[1]{Ruiyuan Jiang}[]
	\ead{Ruiyuan.Jiang20@student.xjtlu.edu.cn}

        \author[1]{Dongyao Jia}[]
	\ead{Dongyao.Jia@xjtlu.edu.cn}

        \author[1]{Eng Gee Lim}[]
	\ead{enggee.lim@xjtlu.edu.cn}

        \author[1]{Pengfei Fan}[]
	\ead{Pengfei.Fan22@student.xjtlu.edu.cn}
 
        \author[1]{Yuli Zhang}[]
	\ead{Yuli.Zhang20@student.xjtlu.edu.cn}
        
        \author[]{Shangbo Wang \textsuperscript{b}\text{\textsuperscript{*}}} 
	\ead{shangbo.wang@sussex.ac.uk}

	\affiliation[1]{organization={School of Advanced Technology},
		addressline={Xi'an Jiaotong-Liverpool University}, 
		city={Suzhou},
		postcode={215123}, 
		country={China}}

        \affiliation[2]{organization={Department of Engineering and Design},
		addressline={University of Sussex}, 
		city={Brighton},
		postcode={BN1 9RH},
		country={UK}}

	\cortext[cor1]{Corresponding authors: Shangbo Wang\textsuperscript{b}\text{\textsuperscript{*}}}

	% Here goes the abstract
	\newcommand{\figref}[1]{Fig.~\ref{#1}}

\begin{abstract}
Accurate traffic prediction is essential for Intelligent Transportation Systems (ITS), yet current methods struggle with the inherent complexity and non-linearity of traffic dynamics, making it difficult to integrate spatial and temporal characteristics. Furthermore, existing approaches use static techniques to address non-stationary and anomalous historical data, which limits adaptability and undermines data smoothing. To overcome these challenges, we propose the Graph Fusion Enhanced Network (GFEN), an innovative framework for network-level traffic speed prediction. GFEN introduces a novel topological spatiotemporal graph fusion technique that meticulously extracts and merges spatial and temporal correlations from both data distribution and network topology using trainable methods, enabling the modeling of multi-scale spatiotemporal features. Additionally, GFEN employs a hybrid methodology combining a \emph{k}-th order difference-based mathematical framework with an attention-based deep learning structure to adaptively smooth historical observations and dynamically mitigate data anomalies and non-stationarity. Extensive experiments demonstrate that GFEN surpasses state-of-the-art methods by approximately 6.3$\%$ in prediction accuracy and exhibits convergence rates nearly twice as fast as recent hybrid models, confirming its superior performance and potential to significantly enhance traffic prediction system efficiency.  
\end{abstract}


\maketitle

\section{Introduction}
In recent years, Intelligent Transportation Systems (ITS) have been increasingly utilized to manage vehicular traffic dynamically and accurately, addressing the growing complexities of traffic demands. As a crucial component of ITS, efficient traffic prediction offers real-time insights for a multitude of practical applications, such as administrative bureaus and navigation systems, by providing reliable data \citep{2015Long}. This capability has the potential to alleviate traffic congestion and enhance safety, thereby optimizing the effective utilization of road network capacity. However, the escalating number of vehicles, particularly during peak hours, has led to frequent traffic congestion. This complex traffic environment results in irregular traffic state data, making the distribution of traffic data irregular and potentially introducing anomalous data into historical observations \citep{YU2025113182, 7065039}. This presents a significant challenge for accurate traffic prediction \citep{7504474}.

Within the realm of existing traffic prediction models, model-driven methodologies such as Autoregressive Integrated Moving Average (ARIMA) \citep{1970Distribution} , Seasonal-ARIMA \citep{2013A}, and Support Vector Regression (SVR) \citep{2003Travel} forecast future traffic scenarios by harnessing the temporal regularity from historical observations, grounded in specific models. However, these model-driven strategies grapple with maintaining high prediction accuracy when confronted with non-linear data structures \citep{9430777}. Conversely, data-driven methodologies, particularly those employing deep learning models, have experienced a significant upswing in popularity, propelled by advancements in computational power. Deep learning models such as Long Short-Term Memory (LSTM) \citep{1997LSTM}, Gate Recurrent Unit (GRU) \citep{2017Discriminative}, T-GCN \citep{8809901}. Specially, with the evolution of computational technology, transformer-based prediction methods have garnered widespread attention \citep{9408777, 2020GMAN}. Their flexibility in network construction and proficiency in extracting spatiotemporal features over extended periods offer distinct advantages, further enhancing their appeal in the field of traffic prediction.

While existing methodologies have achieved remarkable progress in traffic forecasting, two critical research gaps persist in current approaches:

$\bullet$ First, current prediction models face significant limitations in capturing the multi-scale spatiotemporal dependencies inherent in urban traffic networks. Two key shortcomings hinder their effectiveness: 1) Current graph neural architectures insufficiently integrate topology-derived spatial correlations with distribution-based temporal patterns, and 2) The interaction between adaptive graph learning and temporal convolutions remains poorly understood, restricting the model’s ability to extract and leverage multi-scale relationships.

\begin{figure}[h]
	\centering
	\includegraphics[width=0.6\textwidth]{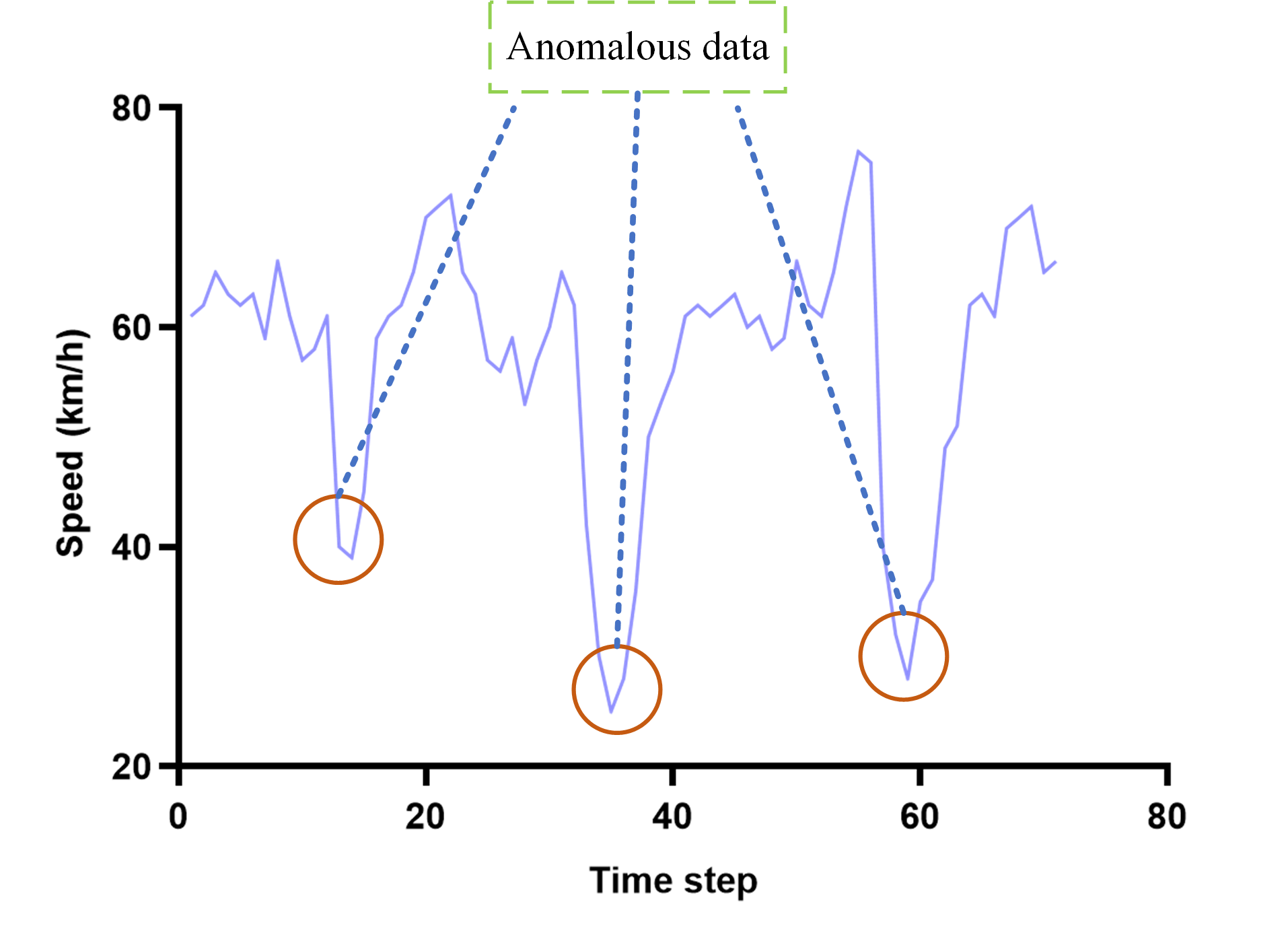}
	\caption{An example of anomalous traffic data.}
	\label{fig:1}
\end{figure}

$\bullet$ Second, traffic data exhibit inherent instability and anomalies that compromise model performance. As visualized in Figure \ref{fig:1}, traffic anomalies manifest as irregular deviations at specific spatiotemporal coordinates \citep{9489378}, typically induced by stochastic events, including accidents, meteorological disruptions, or data acquisition errors. Contemporary solutions employing spectral filtering, wavelet decomposition, and statistical normalization techniques have demonstrated partial success in data rectification. However, these approaches often focus on smoothing or eliminating anomalous data during the preprocessing stage, primarily addressing the problem from a data distribution perspective instead of incorporating the traffic features into the modeling of smooth model. Moreover, these methods usually use mathematical methods with static parameters that cannot adapt during training, which reduces the adaptivity of the methods.

To enhance prediction accuracy and efficiency, we propose the Graph Fusion Enhanced Network (GFEN), implemented in two distinct stages. First, we design an Enhanced Data Correlation (EDC) approach to address the non-stationary and anomalous aspects of traffic data. Second, based on the smoothed data, we introduce a topological spatiotemporal graph fusion technique designed to comprehensively extract spatiotemporal features from both network topology and traffic conditions.

More specifically, our main contributions are summarized as follows:

(1) We introduce a novel multi-graph fusion structure that uses a learnable graph to effectively extract the spatiotemporal correlations of traffic conditions. This innovative approach synthetically combines the spatiotemporal graphs derived from the cross-correlation of traffic data with essential traffic topological information. By leveraging this multi-scale fusion graph within our hybrid prediction model, we improve the modeling of traffic features from different scales, enabling more accurate predictions and deeper insights into traffic dynamics.

(2) To address the adverse effects of anomalous traffic data in historical observations, we propose an Enhanced Data Correction (EDC) technique. This method employs a hybrid mathematical-transformer architecture to adaptively identify and mitigate anomalies while preserving the underlying spatial feature of traffic states.

(3) Extensive experiments indicate that GFEN can achieve higher prediction accuracy with less computational time compared with state-of-the-art methods.

The rest of this paper is structured as follows: Section \ref{sec-lab} reviews existing traffic prediction methods and data smoothing techniques. Section \ref{sec-pre} defines spatiotemporal correlations and outlines the research objectives. In Section \ref{sec-gra}, we introduce the main methodology of the GFEN network. Section \ref{sec-exp} describes the dataset and experimental settings. Section \ref{sec-res} discusses the prediction performance of the GFEN network and compares our model with state-of-the-art methods. Finally, Section \ref{sec-con} and Section \ref{sec-ack} conclude the paper and acknowledge the support for our study.

\section{Related Work} \label{sec-lab}

\subsection{Traffic prediction models}
Over the past decades, traffic prediction has emerged as a prominent research area which is extensively implemented within ITS \citep{7407622, 10490249, 2020Privacy}. Model-driven approaches such as ARIMA \citep{1970Distribution} and its derivatives such as Gaussian ARIMA \citep{9695933} and STARIMA \citep{8525272} use predetermined model structures based on certain theoretical assumptions to extract the temporal factors of time series. Besides, historical average (HA) \citep{2005Predicting} and Kalman Filters \citep{5766715} also raised great attention in the early days.

Apart from the model-driven approaches, data-driven approaches have been widely used for their strong abilities in dealing with stochastic, indeterministic, and non-linearity of traffic data. Various time series prediction approaches such as artificial neural network (ANN) \citep{2013Short}, recurrent unit network (RNN) \citep{9775808}, Bidirectional LSTM (Bi-LSTM) \citep{9332074} and D-GRU \citep{10056481} have made great contributions in forecasting time series. Similarly, spatial information modeling approaches such as GCN-based networks \citep{9565365, 9745164} utilize graph convolutional operation to extract spatial features of traffic states.

Recent research has sought to capture the spatiotemporal correlations in traffic data through graph-based architectures and model fusion techniques \cite{ZHANG2025112966}. For example, Liu \emph{et al.} \citep{LIU2022109760} proposed a hybrid framework integrating a Fuzzy Inference System (FIS) with a Gated Recurrent Unit (GRU), where the GRU extracts temporal patterns while the FIS mitigates the impact of data uncertainty during training. Similarly, Chen \emph{et al.} \citep{CHEN2025130117} proposed a novel approach where historical observations were classified into stationary and non-stationary components, each processed separately for improved prediction. Additionally, they decomposed the topological graph into multiple orders to better characterize road network connectivity, including input-output relationships between different sections. In addition, Cui \emph{et al.} \citep{8917706} designed a TGC-LSTM model to analyze interactions among road segments, incorporating a physical network-based GCN and exploring the relationship between network topology and spectral graph analysis. Zhu \emph{et al.} \citep{9363197} enhanced spatiotemporal graph modeling by introducing an attributed-argument unit that fuses dynamic and static features, leveraging a Seq2Seq-GCN architecture to improve multi-step prediction accuracy. The model extracted the spatiotemporal dependencies through the Seq2Seq model and graph convolutional network, aiming at overcoming the difficulty in the multi-step prediction process. Additionally, Ma \emph{et al.} \citep{2021Short} proposed an LSTM-BiLSTM framework that first analyzes traffic data before transforming complex spatiotemporal relationships into time-series correlation learning. Wang \emph{et al.} \citep{WANG2025129280} designed a comprehensive framework that simultaneously addresses temporal and spatial correlations in traffic data. Their approach employs: (1) a hybrid model integrating convolutional structures with Res2Net to capture temporal dependencies, and (2) a Geom-GCN network to extract global spatial relationships. These distinct correlation features are then dynamically aggregated through an adaptive fusion mechanism to enhance prediction performance. Bai \emph{et al.} \citep{ijgi10070485} further contributed to hybrid modeling approaches. While empirical results demonstrate that these integrated models outperform conventional methods (e.g., standalone LSTM or GRU), they still exhibit limitations in capturing the global dynamics of evolving traffic conditions \citep{pan2019urban}.

Recent studies have demonstrated the superior performance of attention-based architectures in extracting relevant features from global targets \citep{liang2018geoman}. Zhang \emph{et al.} \citep{ZHANG2025130040} a method for topology information extraction by integrating inter-section distance states, connectivity patterns, and graph correlations. Building on this foundation, their framework employs temporal and trend attention mechanisms to capture long-term dependencies, effectively addressing uneven data distribution while modeling traffic dynamics. Li \emph{et al.} \citep{LI2024112381} introduced an attention-enhanced module combined with a dynamically adaptive Graph Convolutional Network (GCN) to model traffic states, incorporating directional, distance-based, and spatiotemporal dependencies. Similarly, Wang \emph{et al.} \citep{10234657} developed an encoder-decoder architecture leveraging multi-head attention and Attention-Gated Recurrent Units (AGRU) to extract multi-scale spatiotemporal features from traffic flow and density data. Moreover, Zheng \emph{et al.} \citep{2020GMAN} proposed GMAN, an encoder-decoder model for traffic flow prediction. Their approach integrates Spatio-Temporal Attention (ST-Attention) blocks to capture dynamic spatial correlations and nonlinear temporal patterns, followed by a fusion mechanism to couple these dependencies. Meanwhile, Jin \emph{et al.} \citep{9713756} designed a Wasserstein GAN (WGAN) framework for urban traffic modeling, employing parallel RNN and GCN branches to learn spatio-temporal representations while optimizing computational efficiency for real-world deployment. Additional innovations include Att-MED by Abdelraouf \emph{et al.} \citep{9535259}, a multi-encoder-decoder model for freeway speed prediction that enhances temporal features through transformer-based attention following LSTM encoding. Yang \emph{et al.} \citep{9564947} adopted a hybrid approach combining attention mechanisms with similarity analysis to identify and incorporate relevant spatial patterns, supplemented by CNN-GRU networks for extracting localized seasonal spatiotemporal features.

\subsection{Data pre-processing methods for historical observations}
In addition to traditional approaches that focus on constructing predictive models, some researchers have adopted pre-processing techniques on historical data prior to the training phase. This strategy aims to minimize the effects of inevitable noise and reduce the influence of anomalous data on prediction accuracy. For instance, Zheng \emph{et al.} \citep{2019DeepSTD} developed the DeepSTD model for citywide traffic prediction, which categorizes historical data based on common area and central building distributions. The DeepSTD model initially removes inherent influencing factors from the historical data and integrates these factors as STD. Subsequently, it employs ResNet to forecast traffic data, excluding STD, and integrates the STD with ResNet's output to enhance prediction accuracy. Ouyang Ouyang \emph{et al.} \citep{OUYANG2023110885} introduced the DAGN model, which extracts spatiotemporal correlations between cities and mitigates domain distribution discrepancies. This model addresses the challenges of training with small data samples in traffic prediction. Additionally, Zhang \emph{et al.} \citep{9348104} utilized mobile edge computing to segregate historical data into trends and regular noise. They processed the trend data using LSTM and predicted random noise using a combination of statistical methods, including Epanechnikov and ARIMA functions, significantly enhancing the prediction accuracy of the raw data. Moreover, \emph{et al.} \citep{saha2023analyzing} proposed a predictive model that incorporates outlier detection with attention mechanisms, demonstrating superior performance over deep sequence models in single-step predictions. Qi \emph{et al.} \citep{qi2024novel} developed a static-based method that leverages diverse information from traffic checkpoint data to filter outlier points in historical travel-time data. This method has proven to be more practical and accurately reflects travel time fluctuations.

Current methodologies have demonstrated remarkable success in addressing data anomalies within prediction models. However, when integrated with deep learning-based prediction frameworks, these approaches introduce a critical drawback: the information loss induced by smoothing operations propagates iteratively during model training, ultimately compromising prediction accuracy. Moreover, most existing pre-processing techniques rely heavily on mathematical assumptions to smooth historical observations based solely on data distribution, failing to account for the inherent spatial characteristics of traffic data.   

\begin{figure}[h]
	\centering
	\includegraphics[width=0.6\textwidth]{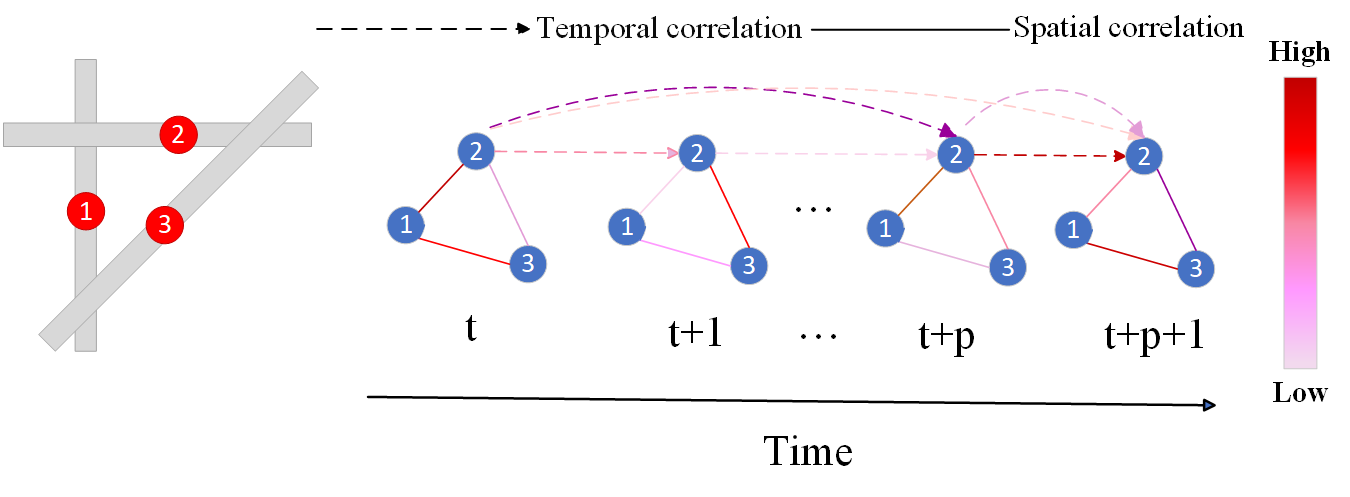}
	\caption{Definitions of spatiotemporal correlations among traffic data}
	\label{fig:2}
\end{figure}
\section{Preliminaries} \label{sec-pre}
To enhance the readability of paper, we outline the following definitions, which have been introduced in \citep{8809901}.

\emph{Definition 1 (Spatial correlation):} Traffic conditions of one section are highly affected by the other sections with different impacts, especially the neighboring road sections and urban hotpots.

\emph{Definition 2 (Temporal correlation):} Traffic conditions at one road section are non-linearly correlated with its historical observations.

\emph{Definition 3 (Spatiotemporal correlations):} Based on \emph{Definition 1} and \emph{Definition 2}, it is evident that the traffic conditions of a section at a specific time step are interconnected not only with its previous observations but also with the traffic conditions of other road sections (see Figure \ref{fig:2}).

\begin{figure}[h]
	\centering
	\includegraphics[width=0.3\textwidth]{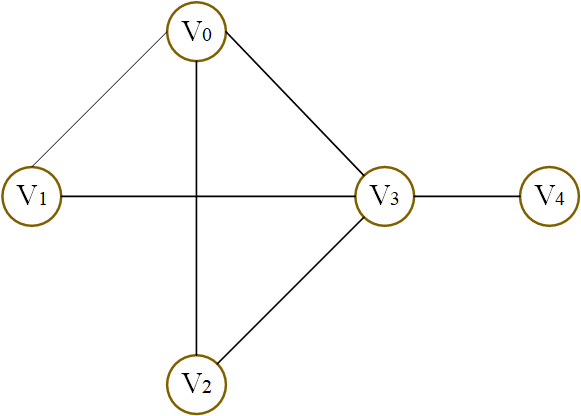}
	\caption{Example of nodes connection structure}
	\label{fig:3}
\end{figure}

\textbf{Problem definition:} A directed graph $\mathcal{G}=(\mathcal{V}, E)$ is applied to represent the topological structure, where $\mathcal{V}$ denotes the set of sensors on roads and $E$ denotes the set of edges reflecting the connectivity among multi-located sensors. Depending on $E$, we apply the adjacency matrix $A\in\mathbb{R}^{N\times{N}}$ to describe the connection states of the road networks. For example, as shown in Figure \ref{fig:3}, where the solid line represents a connection between nodes, and $A$ is described as: 
\begin{equation}
	A=\begin{bmatrix}
		0&1&1&1&0\\
		1&0&0&1&0\\
		1&0&0&1&0\\
		1&1&1&0&1\\
		0&0&0&1&0\\
	\end{bmatrix}
\end{equation}

In this paper, given the historical observations $X^{history}\in\mathbb{R}^{N\times{T}}$ of the previous $T$ time steps and the adjacency matrix $A\in\mathbb{R}^{N\times{N}}$ ($N$ represents the number of sensors), we aim to predict the traffic data $X^{pre}\in\mathbb{R}^{N\times{P}}$ of the next $P$ time steps, shown as:

\begin{equation}
	X_{P}^{pre} = \mathcal{F}(X_{T}^{history})
\end{equation}
where $\mathcal{F}$ represents the model function.

\section {Graph Fusion Enhanced Network} \label{sec-gra}
\begin{figure}[h]
	\centering
	\includegraphics[width=0.8\textwidth]{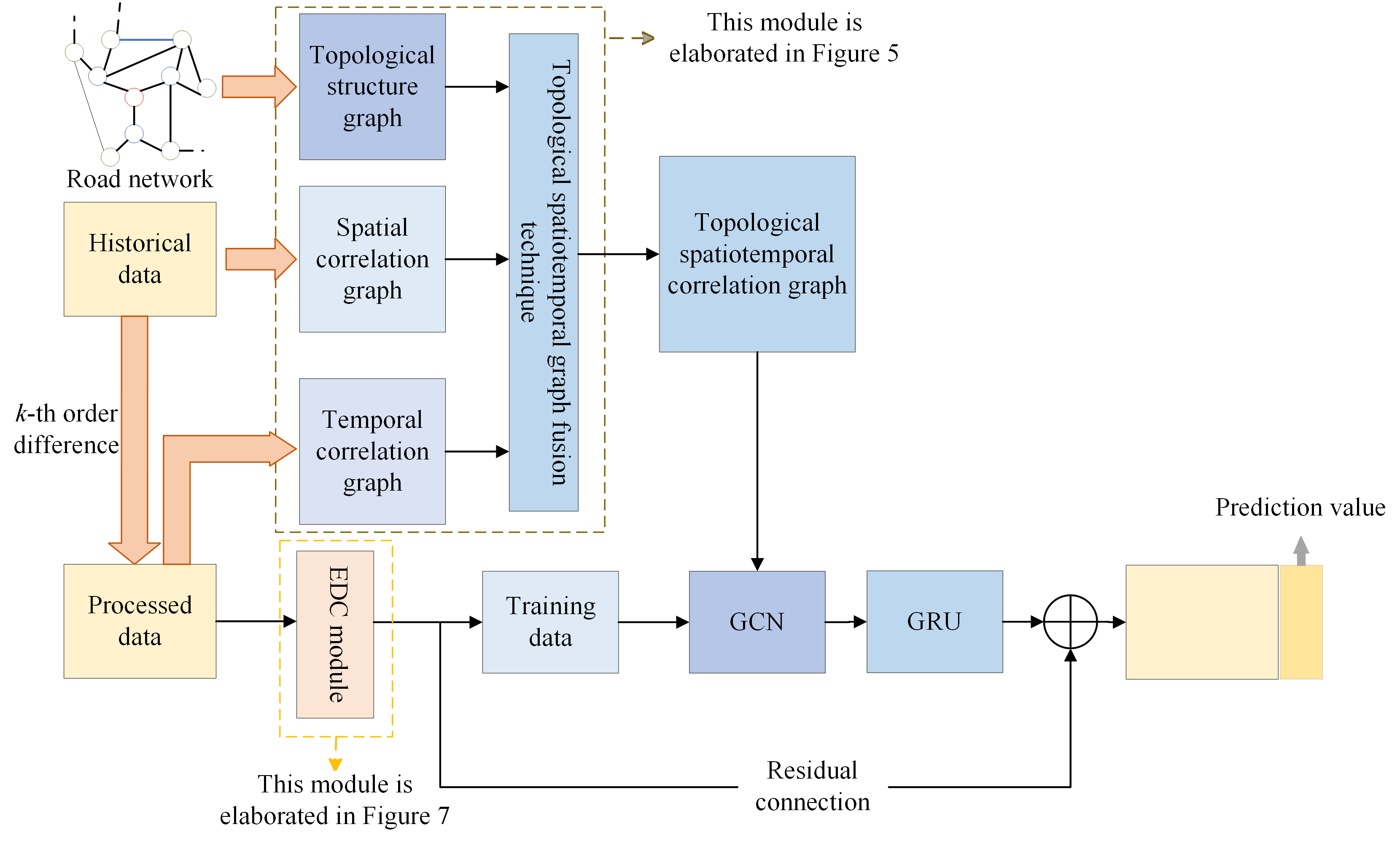}
	\caption{GFEN structure. First, the \emph{k}-th order difference as well as the EDC module is used to mitigate the non-stationarity of historical data. Second, the hybrid model of GCN and GRU is introduced to predict traffic data, where the topological spatiotemporal graph serves as the graph information of the GCN network.}
	\label{fig:4}
\end{figure}

In this study, we propose a Graph Fusion Enhanced Network (GFEN) for predicting network-scale traffic speed. As shown in Figure \ref{fig:4}, we first introduce a topological spatiotemporal graph fusion technique (described in Section \ref{sec-top}) to effectively integrate spatiotemporal correlations into the network topology. Second, we employ a \emph{k}-th order difference approach along with the Enhanced Data Correlation (EDC) module (described in Section \ref{sec-enh}) to smooth the time series data and eliminate anomalies in the traffic data. During the model training process, hybrid models combining GCN and GRU are utilized to extract relevant spatiotemporal features. Each module will be discussed in detail in the subsequent sections.

\subsection{Topological Spatiotemporal Graph Fusion Technique} \label{sec-top}
\begin{figure*}[h]
	\centering
	\includegraphics[width=0.95\textwidth]{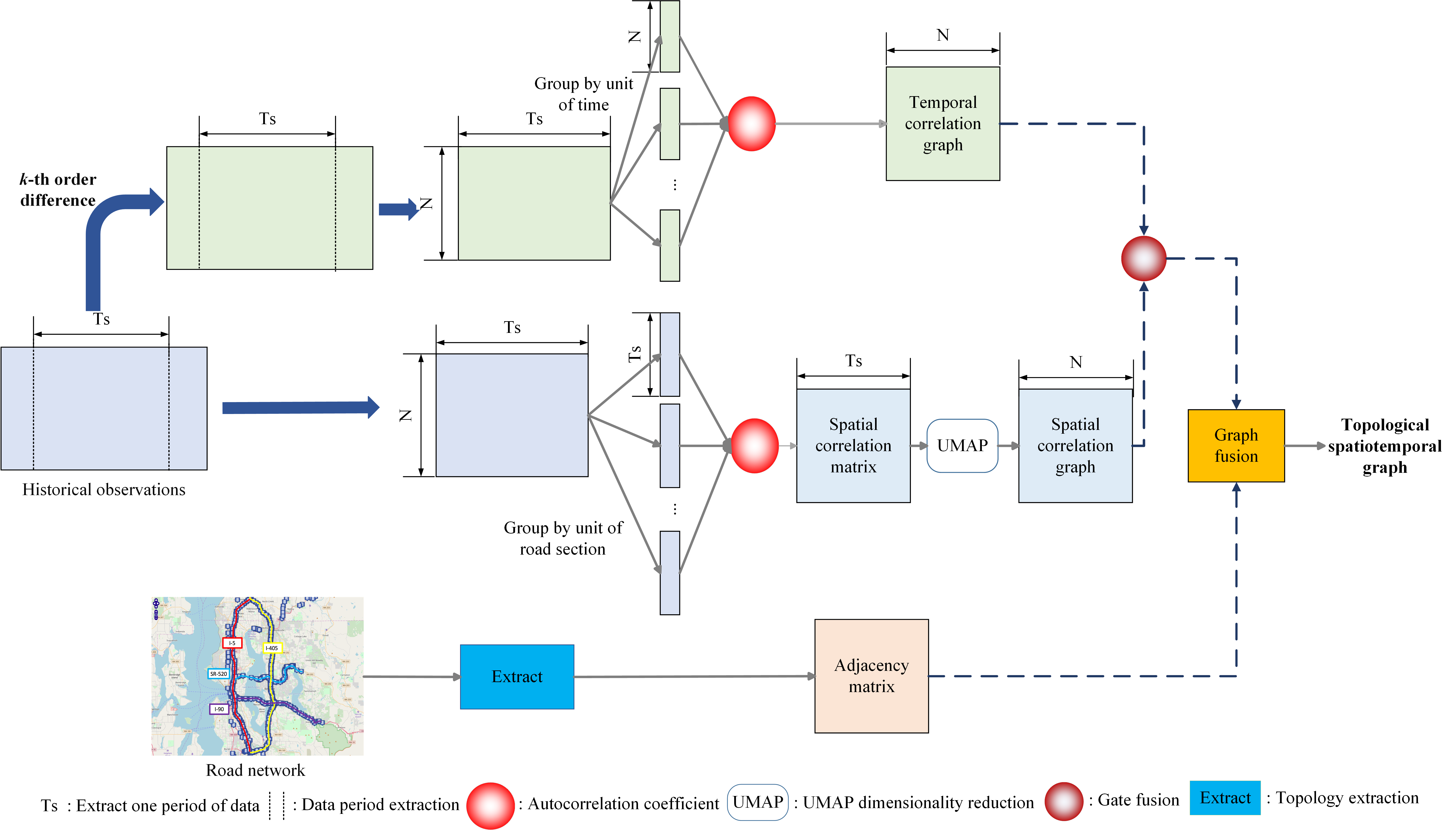}
	\caption{Framework of the topological spatiotemporal graph fusion technique. We first generate the two graphs from the historical data and the data after the \emph{k}-th order difference, respectively. Further, the two graphs and the adjacency matrix are fused into a topological spatiotemporal graph by employing the two-step graph fusion technique.}
	\label{fig:5}
\end{figure*}

To capture the spatiotemporal features of traffic data at a deeper level, we propose a Topological Spatiotemporal Graph Fusion (TSTGF) technique. This technique facilitates the generation of a hybrid graph model that effectively integrates spatial correlations, temporal correlations, and the overall global network structure. As illustrated in Figure \ref{fig:5}, we first create temporal and spatial graphs from historical observations, and then construct a topological graph based on the road network structure. The final topological spatiotemporal graph is generated by fusing these three graphs through a dedicated graph fusion module. The framework of the TSTGF technique is detailed in the following sections.

\subsubsection{Periodic Analysis} Prior to delving into the spatiotemporal correlations, we initially undertake the extraction of a single period's worth of data for the purpose of correlation modeling. This step is crucial for mitigating the excessive computational demands that arise from analyzing the entirety of historical observations. Utilizing the processed historical observations ($\tilde{X}$), we embark on a periodic analysis of traffic data spanning the entire network. The insights gleaned from this analysis will be further elaborated upon in Section \ref{sec-top}. To facilitate this comprehensive periodic analysis, we aggregate speed data from all road segments, forming the basis of our investigation:
\begin{equation}
	X^{sum} = \sum_{j=1}^{N}{\tilde{X}_{j}}
\end{equation}
In our quest to pinpoint the periodic intervals within the dataset with greater precision, we estimate the traffic data's period by identifying peaks within its frequency domain. This involves the application of the Fourier Transform to transpose the traffic data into the frequency domain. For the time series ($X^{sum}(t)$), the Fourier Transform is articulated as follows:
\begin{equation}
	F(\omega)=\mathcal{F}(X^{sum}(t))=\int_{-\infty}^{+\infty}X^{sum}(t)e^{-i\frac{2\pi}{T}t}dt
\end{equation}
where $F(\omega)$ is the form of $X^{sum}$ in the frequency domain, $\mathcal{F}$ represents the Fourier Transform operation, $T$ denotes the sampling period, and $i$ represents the imaginary unit. Based on $F(\omega)$, we obtain the period of historical observations by observing the peaks and we define it as $T_{s}$.

\subsubsection{Spatiotemporal Features Modeling}
Building upon the single period of historical observations obtained in the previous step, we proceed to extract spatiotemporal features. The steps for this process are outlined as follows.

\textbf{Spatial correlation block} Based on the original historical observations $X$, we aim to generate the spatial correlation graph by calculating the spatial correlation among sensors. 
For traffic data of a single sensor, we consider it as a one-dimensional time series and define it as ${X_{t}, t\in(0, T-1)}$. According to $X_{t}$, we define the mean function of this series as:
\begin{equation}
	\mu_{t}=E[X_{t}]=\frac{\sum_{t=0}^{T-1}X_{t}}{T}
\label{eq.5}
\end{equation}
where $\mu_{t}$ represents the corresponding mean function sequence of $X_{t}$. Therefore, according to the mean function, the variance of $X_{t}$ can be defined as follows: 
\begin{equation}
	\sigma_{t}=E(X_{t}-\mu_{t})^2=\frac{1}{T}\sum_{t=0}^{T-1}(X_{t}-\mu_{t})^2
\label{eq.6}
\end{equation}\\
Let $X^{m}$ be the traffic data at the \emph{m}-th sensor and $X^{n}$ be the traffic data at the \emph{n}-th sensor. Based on the definitions of mean function and variance, the spatial correlation between the corresponding two sensors can be defined as:
\begin{equation}
	Spcor(m, n)=\frac{E[(X^{m}_{t}-\mu^{m}_{t})(X^{n}_{t}-\mu^{n}_{t})]}{\sqrt{\sigma^{m}_{t}\sigma^{n}_{t}}}
\end{equation}
as the spatial correlation coefficient between sensor $m$ and sensor $n$, where $X^{m}_{t}$ and $X^{n}_{t}$ represents the time series of the two sensors, $\mu^{m}_{t}$ and $\mu^{n}_{t}$ denotes the mean function of the series, and $\sigma^{m}_{t}$ and $\sigma^{n}_{t}$ are the variance of the time series.

To reduce the computational time, for each sensor, we extract the data within a period for spatial correlation calculation, and the spatial correlation coefficient between sensor $m$ and sensor $n$ can calculated as:
\begin{equation}
	\begin{split}
		S(m, n)=\frac{E[(X^{m}_{t}[c: c+T_{s}]-\mu^{m}_{t}[c: c+T_{s}])
		(X^{n}_{t}[c, c:T_{s}]-\mu^{n}_{t}[c: c+T_{s}])]}{\sqrt{\sigma^{m}_{t}[c: c+T_{s}]\sigma^{n}_{t}[c: c+T_{s}]}}
	\end{split}
\label{eq.8}
\end{equation}
where $T_s$ is the period of traffic data which can be obtained from the process of periodic analysis in Section \ref{sec-tra}, $c\in(1, T-T_{s})$ is a constant, $X^{m}_{t}[c: c+T_{s}]$ represents the data in one period of the time series collected from the sensor $m$, $\mu^{m}_{t}[c: c+T_{s}]$ and $\sigma^{m}_{t}[c: c+T_{s}]$ are the mean function and variance of the data for the period, which are the same for sensor $n$.

From Eq.\ref{eq.8}, the spatial correlation graph can be generated as:
\begin{footnotesize}
	\begin{equation}
		\hat{GS}=\begin{bmatrix}
			S(1, 1)&S(1, 2)&\ldots&S(1, N)\\
			S(2, 1)&S(2, 2)&\ldots&S(2, N)\\
			S(3, 1)&S(3, 2)&\ldots&S(3, N)\\
			\vdots&\vdots&\ddots&\vdots\\
			S(N-1, 1)&S(N-1, 2)&\ldots&S(N-1, N)\\
			S(N, 1)&S(N, 2)&\ldots&S(N, N)\\
		\end{bmatrix}
	\end{equation}
\end{footnotesize}
\begin{equation}
	GS=softmax(\hat{GS})
\end{equation}
where $GS$ denotes the spatial correlation graph, $softmax(\cdot)$ denotes the softmax function which makes the graph non-linear.

\textbf{Temporal correlation graph} The traffic condition of a certain time step can be affected by its previous observations. Therefore, we generate a temporal correlation graph to extract the temporal feature. For the processed data $\tilde{X}$, we calculate the temporal correlation between the traffic data of each time step. To reduce the computation time, we select one period of historical data for correlation coefficient calculation, so the computation complexity can be reduced from $T^2$ to $T_{s}^2$ ($T_{s}$ is the length of one period which is obtained in the previous stage). Similar to the definition of mean function and variance of one-dimension series which are shown in Eq.\ref{eq.5} and Eq.\ref{eq.6}, the temporal correlation between time step $p$ and $q$ ($p, q\in(c, c+T_{s})$, $c\in(0, T-T_{s})$ is a constant) can be defined as:
\begin{equation}
	T(p, q)=\frac{E[(\tilde{X}_{p}-\mu_{p})(\tilde{X}_{q}-\mu_{q}])}{\sqrt{\sigma_{p}\sigma_{q}}}
\label{eq.11}
\end{equation}
where $T(p, q)$ is the temporal correlation coefficient, $\tilde{X}_{p}\in\mathbb{R}^{N\times 1}$ and $\tilde{X}_{p}\in\mathbb{R}^{N\times 1}$ are the data series for all sensors of time step $p$ and $q$ in this period, $\mu_{p}$ and $\mu_{q}$ are the mean functions of the two series, $\sigma_{p}$ and $\sigma_{q}$ are the variances of the two series.

Based on Eq.\ref{eq.11}, the temporal correlation graph can be generated as:
\begin{footnotesize}
	\begin{equation}
		\hat{GT}=\begin{bmatrix}
			T(1, 1)&T(1, 2)&\ldots&T(1, T_{s})\\
			T(2, 1)&T(2, 2)&\ldots&T(2, T_{s})\\
			T(3, 1)&T(3, 2)&\ldots&T(3, T_{s})\\
			\vdots&\vdots&\ddots&\vdots\\
			T(T_{s}-1, 1)&T(T_{s}-1, 2)&\ldots&T(T_{s}-1, T_{s})\\
			T(T_{s}, 1)&T(T_{s}, 2)&\ldots&T(T_{s}, T_{s})\\
		\end{bmatrix}
        \label{eq.12}
	\end{equation}
\end{footnotesize}
where $\hat{GT}$ denotes the original temporal correlation graph. To facilitate the process of graph fusion, the dimensions of the temporal correlation graph and the spatial correlation graph should be the same. Hence, we reduce the dimension of the time correlation graph which is obtained in Eq.\ref{eq.12}. With consideration of the non-linearity of the temporal correlation of traffic data, we apply the uniform manifold approximation and projection (UMAP) to reduce the dimension of $\hat{GT}$.

UMAP retains the global structure of the original data \citep{2021The}, with superior operational skills and good scalability \citep{9585419}. Consequently, many machine learning applications have adopted UMAP as a dimensionality reduction technique. UMAP initially constructs a high-dimensional representation of the data, followed by the optimization of a low-dimensional graph to ensure that its architecture closely resembles that of the original high-dimensional representation. For a detailed explanation of the methodology, please refer to \citep{2018UMAP}. Benefiting from UMAP, we can accurately reduce the dimension of $\hat{GT}$ to the same dimension as spatial correlation graph $GS$:
\begin{equation}
	\tilde{GT} = UMAP(\hat{GT})
\end{equation}
where $\tilde{GT}$ represents the temporal correlation graph after dimensionality reduction, $UMAP(\cdot)$ denotes the process of UMAP. We apply the \emph{softmax} function to non-linearize the graph $\tilde{GT}$.
\begin{equation}
	GT = softmax(\tilde{GT})
\end{equation}
where $UMAP(\cdot)$ denotes the process of UMAP, $GT$ represents the final temporal correlation graph.

\textbf{Graph fusion} Based on the spatial correlation graph, temporal correlation graph, and adjacency matrix, we design a graph fusion model to synthesize the information of the three graphs into one graph. We first use the gated fusion technique to incorporate the spatial and temporal representations. As shown in Figure \ref{fig:6}, the spatial correlation graph and temporal correlation graph are fused as: 
\begin{figure}[h]
	\centering
	\includegraphics[width=0.35\textwidth]{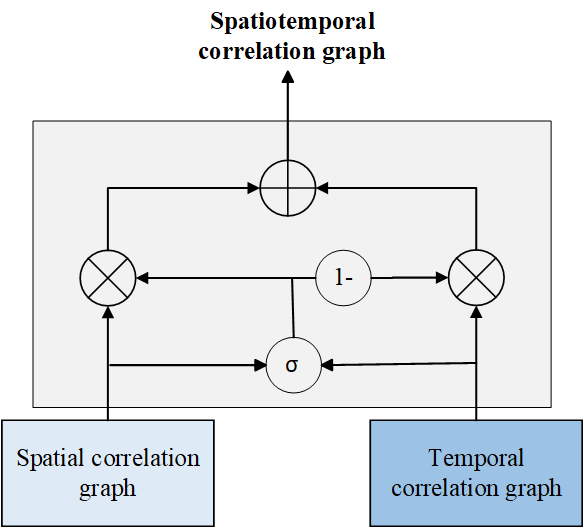}
	\caption{Gate fusion structure}
	\label{fig:6}
\end{figure}
\begin{equation}
	GST = z\odot{GS}+(1-z)\odot{GT}
\label{eq.15}
\end{equation}
with
\begin{equation}
	z = \sigma(GSW_{z, 1}+GTW_{z, 2}+b_{z})
\end{equation}
where $GST\in\mathbb{R}^{N\times{N}}$ represents the spatiotemporal graph, $W_{z, 1}$, $W_{z, 2}$ and $b_{z}$ are learnable parameters, $\odot$ denotes the element-wise product, $\sigma(\cdot)$ denotes the sigmoid function. The gated fusion adaptively controls the spatiotemporal dependencies of traffic information at each sensor and time step.

To consider the spatiotemporal dependencies synthetically with the connection situation between each road section, we propose a topological spatiotemporal graph to fuse the spatiotemporal graph $GST$ generated in Eq.\ref{eq.15} and the topological structure graph. In this graph, when there is a connection between the two sections, we set the correlation coefficient of the corresponding position in the spatiotemporal graph to the value of that point. And if there is no connection, we set the value of that point as 0. Specifically, the graph fusion process can be defined by: \\
$\forall i\in(1, N), j\in(1, N)$
\begin{equation}
	\begin{split}
		TG_{i, j} = \begin{cases}
			GST_{i, j}, & A_{i, j}=1\\
			0, & A_{i, j}=0
		\end{cases}
	\end{split}
\end{equation}
where $TG_{i, j}$ represents the $i$-th column of the $j$-th raw in topological spatiotemporal graph $TG$.

Specifically, utilizing the topological spatiotemporal graph fusion technique, we employ the topological spatiotemporal graph $TG$ to comprehensively extract spatiotemporal correlations from both traffic data and network structures. This approach enhances the efficiency of coupling spatiotemporal correlations during model training.

\subsection{Enhanced Data Correlation} \label{sec-enh}
As previously noted, the historical traffic state observations may exhibit non-stationarity and data anomalies due to the complex traffic environment, which can negatively impact prediction accuracy. To address this challenge, we introduce an Enhanced Data Correlation (EDC) method in this paper. This method is a hybrid framework that combines an order difference-based mathematical method with an attention-based transformer structure. The primary objective of the EDC method is to adaptively smooth historical observations while taking into account the potential spatial features of traffic states. The specifics of each structure within this method will be detailed in the following sections.

\subsubsection{Mathematical-based Approach for Traffic Data Smoothing} \label{sec-tra}
\textbf{Data stationarity} In this paper, we apply the \emph{k}-th order difference as the mathematical approach to mitigate the non-stationarity. Specifically, let us express historical traffic data as $X=[X^1, X^2, ..., X^{N}]$, where $X^{j}\in\mathbb{R}^{1\times{T}}$ denotes the data of the j-\emph{th} sensor. The \emph{k}-th order difference is defined as: 
\begin{equation}
	\nabla_{k}X_{t}^{j} = X_{t}^{j}-X_{t-k}^{j}
\end{equation}
where $t$ is the number of terms, and \emph{k} is the order of difference. Specifically, the first-order difference can be defined as:
\begin{equation}
	\nabla X_{t}^{j} = X_{t}^{j}-X_{t-1}^{j}
\end{equation}
We define $\tilde{X}\in\mathbb{R}^{N\times{T-k}}$ as the processed historical observations after \emph{k}-th order difference operation for traffic data of each sensor. Here, we consider each series in $\tilde{X}$ to be a wide-stationary time series. Then, we employ $\tilde{X}$ as the input, which aims to improve the efficiency of periodic extraction and estimation of model parameters in the following stages.

\subsubsection{Transformer-based Approach for Traffic Data Smoothing} \label{sec-enh}
\begin{figure*}[t]
	\centering
	\includegraphics[width=1\textwidth]{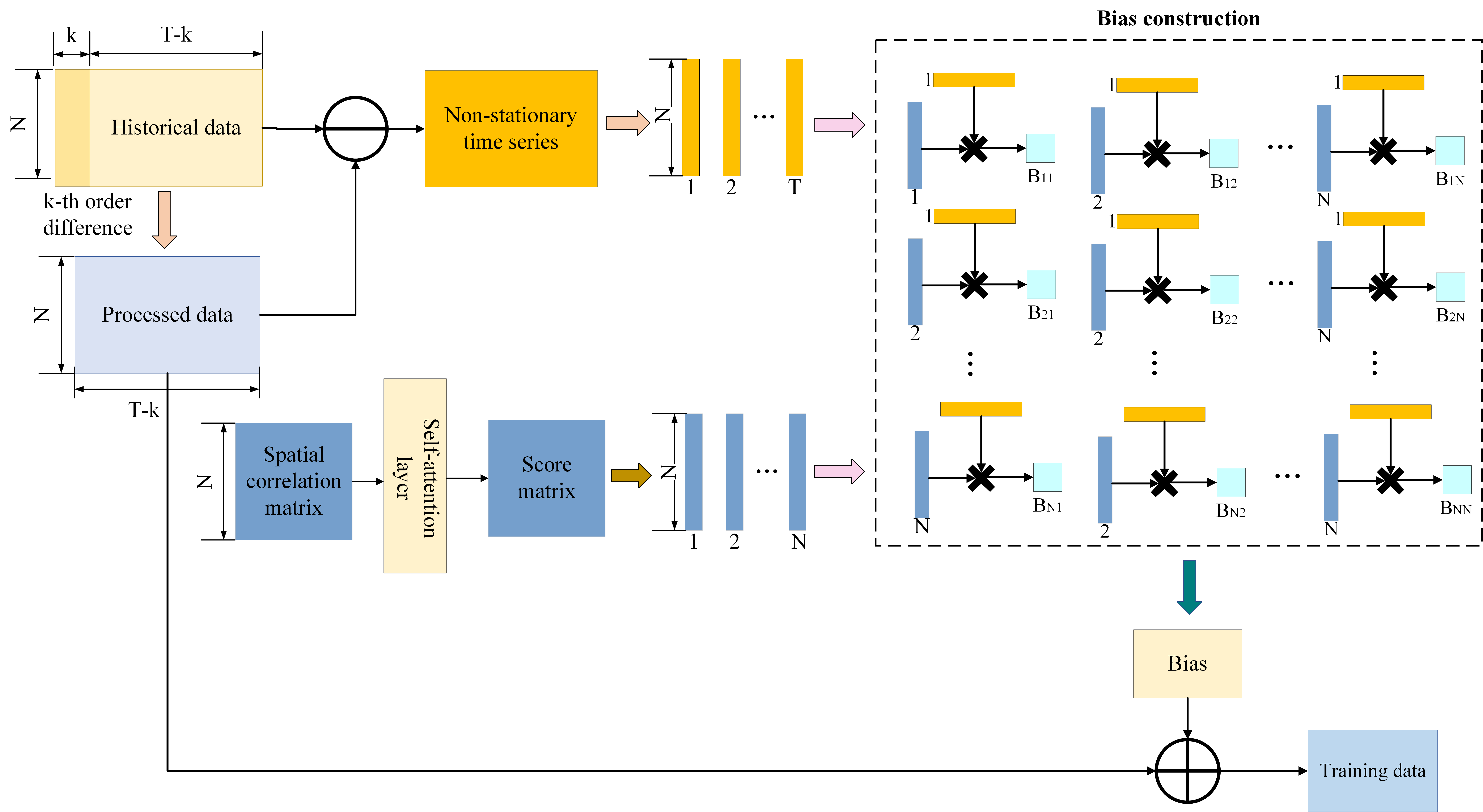}
	\caption{Structure of the Enhanced Data Correlation}
	\label{fig:7}
\end{figure*}

While we employ the \emph{k}-th order difference in the preceding phase (Section \ref{sec-tra}) to initially smooth the historical observations, it proves challenging to mitigate points exhibiting significant volatility. To diminish the adverse effects of outlier data, we introduce a refined Enhanced Data Correlation (EDC) methodology that leverages a transformer to adaptively attenuate these fluctuations during model training. The primary objective of this architecture is to incorporate a trainable bias into the historical data, with the goal of normalizing the outliers. The formulation of this bias is entirely predicated on the historical data and the spatial feature among the traffic data across different road segments. As depicted in Figure \ref{fig:7}, our initial step involves isolating points that demonstrate a stationary distribution offset. Utilizing the original dataset $X$ and the modified dataset $\tilde{X}$, derived as per Equation (3), we delineate the set of non-stationary time series as follows:
\begin{equation}
	\hat{X} = X[k, T]-\tilde{X}
\end{equation}
where $k$ represents the order of \emph{k}-th order difference, $X[k, T]$ denotes the data after the \emph{k}-th time steps of $X$. 

In this context, our objective is to construct a bias that will decrease the values in $\hat{X}$, thereby mitigating the impact of anomalous data. Leveraging the spatial correlation graph $GS$, we utilize self-attention to further compute the dynamic correlation coefficient. It is well understood that the spatial correlation of traffic data is highly variable, fluctuating across different time steps. Consequently, we dynamically assign varying weights to different sensors, enabling an adaptive adjustment of the correlation matrix values and providing an accurate depiction of spatial correlation at distinct time steps. More specifically, we employ the scaled dot-product attention method \citep{NIPS2017_3f5ee243} to generate the score matrix:
\begin{equation}
	SC = softmax(\frac{\left \langle W_{q}GS, W_{k}GS \right \rangle}{\sqrt{N}})\cdot{W_{v}GS}
\end{equation}
where $SC$ $\left \langle \cdot \right \rangle$ represents the inner product, $W_{q}$, $W_{k}$ and $W_{v}$ are learnable parameters to generate query, key, and value of the spatial correlation graph $GS$.

In the score matrix $GS$, each row represents the spatial correlation between the traffic data of the corresponding sensors and those of the other sensors. For example, $GS_{i}$ represents the spatial correlation coefficient between sensor $i$ and the other sensors, and $GS_{i, j}$ denotes the correlation of traffic data between sensor $i$ and sensor $j$.

In the EDC module, we set each value of the bias with the weighted sum of the data of all positions and the spatial correlation coefficients of the corresponding rows in the score matrix $SC$. We first divide $SC$ into $N$ raws which represents the coefficient series for the corresponding $N$ sensors:
\begin{equation}
	SC = [SC_{1}, SC_{2}, ...,SC_{N}]
\end{equation}
We assume that one abnormal pattern happens in the \emph{i}-th sensor at the \emph{j}-th time step, and the value of the corresponding position in the trainable bias can be defined as: 
\begin{equation}
	B_{i, j}=\sum_{h=1}^{N}\hat{X}_{j, N}\cdot{SC_{i, N}}=\hat{X}_{j}\times{SC_{i}}
\end{equation}
Then, the bias can be obtained by:
\begin{equation}
	B=\begin{bmatrix}
		B(1, 1)&\ldots&B(1, T-k)\\
		B(2, 1)&\ldots&B(2, T-k)\\
		\vdots&\ddots&\vdots\\
		B(T-k, 1)&\ldots&B(T-k, T-k)\\
	\end{bmatrix}
\end{equation}
where $\lvert \rvert$ represents the contact of vector, $B_{t}$ denotes the vale of bias at the \emph{t}-th time step.

Based on the trainable bias $B\in\mathbb{R}^{N\times{T-k}}$, the traffic data which serves as the input of model training can be calculated by:
\begin{equation}
	X^{in}=\tilde{X}+B
\end{equation}
Hence, if there is an abnormal pattern, the deviation between this point and the stationary value can be reduced by the normal values of the other sensors at the same time step. So the error in parameter estimation caused by anomalous data generated by several sensors in a one-time step will be distributed among all sensors in that time step, which can reduce the negative impact of these anomalous data during model training. Further, the stationarity of the historical data will improve the efficiency of model training, so the computational time can be reduced during model training.

\subsection{Model Structure}
As shown in Figure \ref{fig:4}, we propose the GFEN network to timely and accurately predict the future traffic data. Based on the topological spatiotemporal graph $TG$ generated by the topological spatiotemporal graph fusion technique (Section \ref{sec-top}) and the training data $X^{in}$ processed by the EDC module (Section \ref{sec-enh}), we first apply the GCN network to model spatial features and couple the temporal features of traffic information. We define the output of one layer of GCN as:
\begin{equation}
	H^{n+1} = \sigma(\widetilde{D}^{-\frac{1}{2}}\widetilde{TG}\widetilde{D}^{-\frac{1}{2}}H^{n}\theta_{n})  
\end{equation}
\begin{equation}
	\widetilde{TG} = TG + I_N  
\end{equation}
where $\widetilde{D}=\sum_{j}{\tilde{TG}}$ is the degree matrix of $TG$, $H^{n+1}$ represents the output of the \emph{n+1}-th layer, $I_N\in\mathbb{R}^{N\times{N}}$ represents an identity matrix, $H^{n}$ is the output of the \emph{n}-th layer, $\theta_{n}$ are the parameters of GCN network. Here, two layers of the GCN are employed and the output of GCN can be obtained by:

\begin{equation}
	\widetilde{S} = (\sum_{j}{\widetilde{TG}})^{-\frac{1}{2}}\widetilde{TG}(\sum_{j}{\widetilde{TG}})^{-\frac{1}{2}}
\end{equation}
\begin{equation}
	O_{G} = \sigma(\widetilde{S}\mathrm{ReLU}(\widetilde{S}X^{in}W_1)W_2)
\end{equation}
where $W_1$ and $W_2$ represent the model parameters, and $ReLU$ denotes the Rectified Linear Activation Function. \par

Next, we utilize a GRU to further extract temporal features, using the output of the GCN as input. The GRU network comprises three main components: the update gate, cell gate, and reset gate. For input data $X_{t}$ at time (t), the states of these gates and the output are described as follows:
\begin{equation}
	u_{t} = \sigma(W_{u}[X_{t}, h_{t-1}]+b_{u})
\end{equation}
\begin{equation}
	r_{t} = \sigma(W_{r}[X_{t}, h_{t-1}]+b_{r})
\end{equation}
\begin{equation}
	c_{t} = tanh(W_{c}[X_{t}, (r_{t}\cdot{h_{t-1}})]+b_{c})
\end{equation}
\begin{equation}
	\hat{Y_{t}} = u_{t}*h_{t-1}+(1-u_{t})*c_{t}
\end{equation}
where $W_{u}$, $W_{r}$, $W_{c}$ and $b_{u}$, $b_{r}$ and $b_{c}$ are the learnable weights and biases for GRU, $u_{t}$,$r_{t}$ and $c_{t}$ denote the state of update gate, reset gate and cell gate at time \emph{t}, $\hat{Y_{t}}$ denotes the output of GRU. 
% \begin{figure}[h]
% 	\centering
% 	\includegraphics[width=0.4\textwidth]{GRU.png}
% 	\caption{Gated recurrent unit neural network structure}
% 	\label{fig:8}
% \end{figure}

In model training, we apply the residual connection \citep{2016Residual} in Eq. \ref{eq.34} to ensure the training stability, shown as:
\begin{equation}
	Y_{t}=W_{out}\hat{Y_{t}}+W_{res}X^{in}
\label{eq.34}
\end{equation}
where $Y_{t}$ denotes the prediction results of GFEN, $W_{out}$ and $W_{res}$ denotes the trainable model parameters.

Our framework begins with a topological spatiotemporal graph fusion module to jointly capture spatial and temporal dependencies in the data. This integrated representation enables the subsequent GRU network to efficiently model temporal dynamics, accelerating model convergence. To further enhance training stability and efficiency, we apply the EDC technique to smooth the historical observations, ensuring stationarity in the input data. This step not only refines parameter estimation accuracy but also reduces computational overhead by optimizing the training process.

During training, the model parameters of GFEN are optimized by minimizing the loss between prediction results and real data through backward propagation. We set the loss function of our model as: 
\begin{equation}
	L = \lvert \lvert Y_{t}-\widehat{Y_{t}} \rvert \rvert +\lambda L_{reg}
\end{equation}
where $Y_{t}$ represents the real traffic data, $\widehat{Y_{t}}$ represents the prediction results, $\lambda$ denotes a hyper parameter, and $L_{reg}$ is a L2 regularization term. 

\section{Experimental Design} \label{sec-exp}
\subsection{Data Description}
We test the GFEN model on two real-world datasets: Los-loop and Seattle-loop datasets.

\begin{figure}[h]
	\centering
	\subfloat[]{
		\includegraphics[scale=0.38]{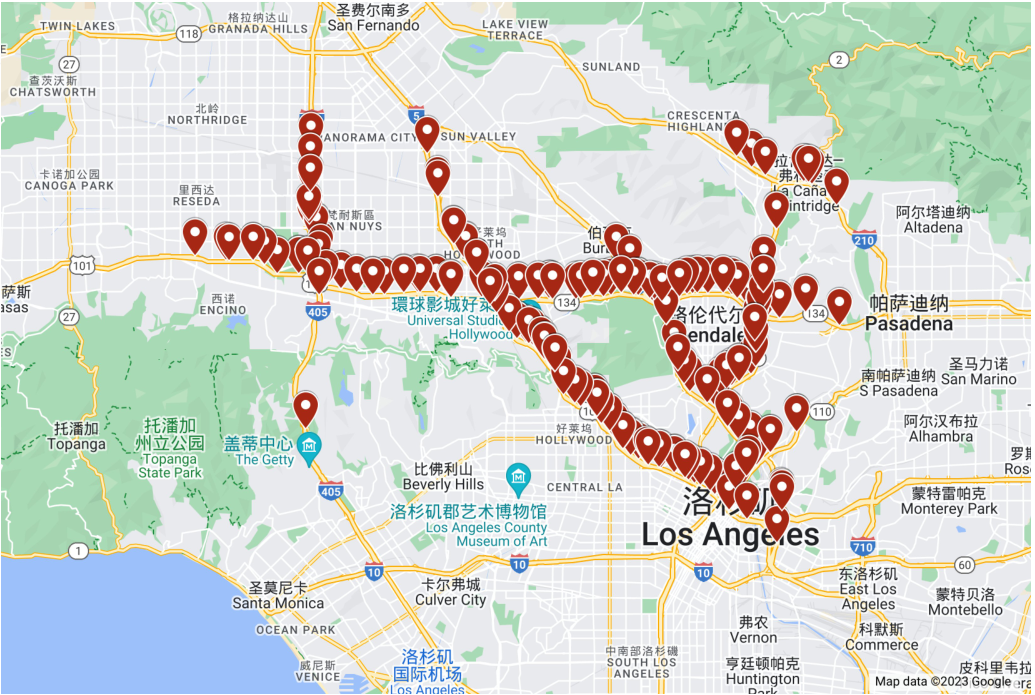}}
	\subfloat[]{
		\includegraphics[scale=0.35]{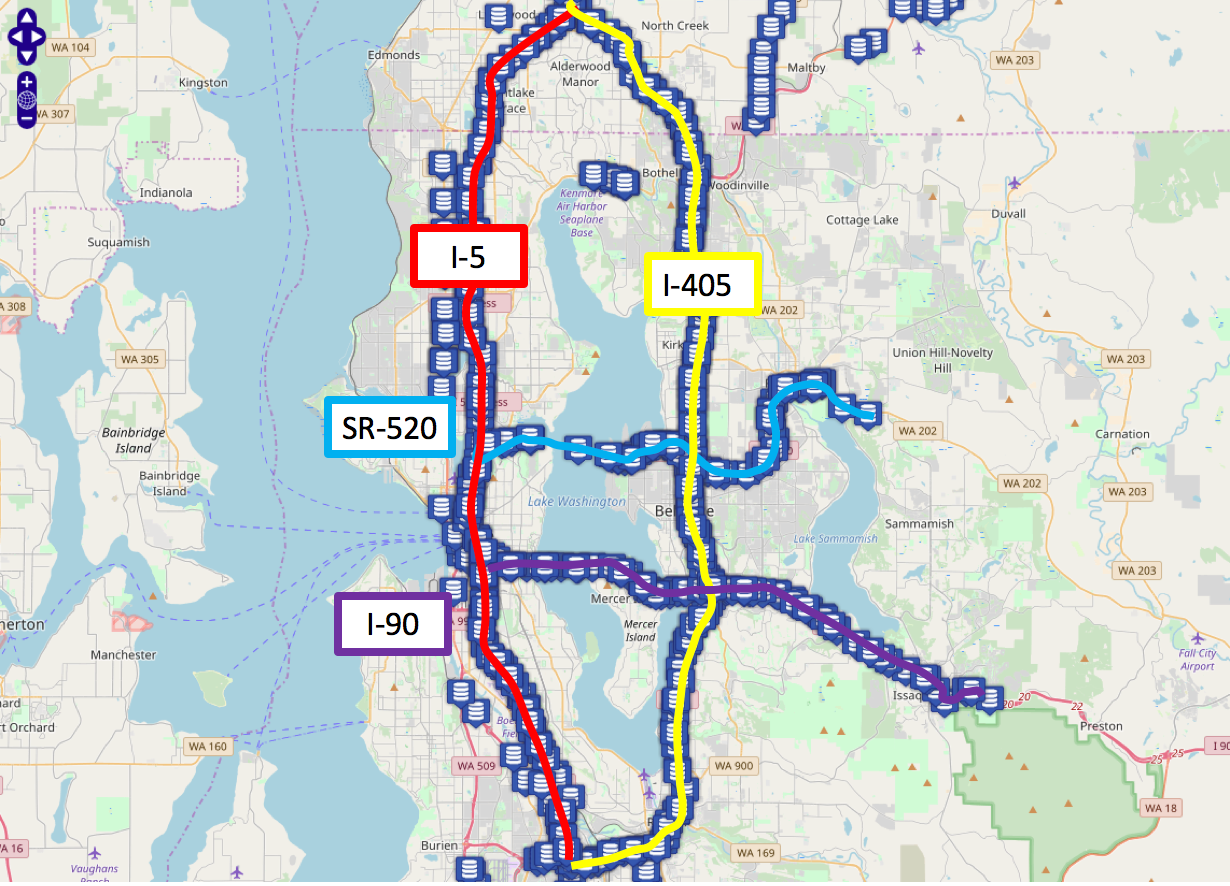}}
	\caption{Locations of sensors in two datasets. (a) Los-loop dataset. (b) Seattle-loop dataset}
	\label{fig:8}
\end{figure}

\textbf{Los-loop dataset} This dataset provides the traffic speed data in Los Angeles country from March 1, 2012, to March 7, 2012. The data are collected from 207 sensors and the sampling time interval is 5 minutes, where the distributions of sensors are shown in Figure \ref{fig:8} (a). The first part of this dataset is a data matrix that consists of traffic speed data of all sensors over all time steps, where each row represents the speed data of one corresponding sensor and each column denotes data from all sensors at one step. The second part is a $207\times{207}$ adjacency matrix representing the connection conditions among all sensors. \par 

\textbf{Seattle-loop dataset} The Seattle-loop dataset describes the traffic speed data from the inductive loop sensors in the Seattle area in 2015 and the distributions of sensors are shown in Figure \ref{fig:8} (b). 
The dataset includes two parts, the first part consists of spatiotemporal speed information of the freeway system, where the speed information at a milepost is averaged from the different sensors on the main stems at the specific milepost. The second part is a loop adjacency matrix which contains the connection states between each pair of sensors.

\subsection{Evaluation Metrics}
The following metrics—RMSE, MAE, $Acc$, $R^2$, and VAR—are employed to evaluate the predictive performance of our GFEN model. These evaluation metrics are essential for quantifying the accuracy and reliability of our model’s predictions compared to the actual observed data. Here, $Y$ symbolizes the true data values, while ($Y^{p}$) represents the predicted outcomes generated by the model. By systematically assessing these metrics, we can gain valuable insights into the model's performance across various dimensions. \\
1) Root Mean Square Error (RMSE)
\begin{equation}
	RMSE = \sqrt{\frac{1}{MN}\cdot\sum_{j=1}^M{\sum_{i=1}^N{{(Y_{i, j}-Y_{i, j}^{p})^2}}}}
\end{equation}
2) Mean Absolute Error (MAE)
\begin{equation}
	MAE = \frac{1}{MN}\cdot\sum_{j=1}^{N}{\lvert Y_{i, j}-Y_{i, j}^{p} \rvert}
\end{equation}
3) Accuracy ($Acc$)
\begin{equation}
	Acc = 1-\frac{\lvert \lvert Y_{i, j}-Y_{i, j}^{p} \rvert \rvert_{F}}{\lvert \lvert Y_{i, j} \rvert \rvert_{F}}
\end{equation}
4) Coefficient of Determination ($R^2$)
\begin{equation}
	R^2 = 1-\frac{1}{MN}\cdot\sum_{j=1}^M{\sum_{i=1}^N{{(Y_{i, j}-Y_{i, j}^{p})^2}}}{(Y_{i, j}-\bar{Y})^2}
\end{equation}
5) Explained Variance Score (VAR)
\begin{equation}
	VAR = 1-\frac{f_{var}(Y_{i, j}-Y_{i, j}^{p}}{f_{var}(Y)}
\end{equation}

\subsection{Baselines}
To evaluate the prediction accuracy and efficiency of GFEN, we select following baseline methods for comparison: (1) \textbf{ARIMA} \citep{1997ARMA}; (2) Historical average (\textbf{HA}) \citep{2004INCORPORATING} (3) Support vector regression (\textbf{SVR}) \citep{2003Travel}; (4) \textbf{LSTM} \citep{1997LSTM}; (5) \textbf{T-GCN}, that combines GCN with GRU network \citep{8809901}; (6) Attention temporal graph convolutional network (\textbf{A3T-GCN}), that combines T-GCN prediction network with attention mechanism to extract global temporal dynamics and spatial correlation \citep{ijgi10070485}; (7) Traffic graph convolutional long short-term memory neural network (\textbf{TGC-LSTM}), which apply traffic graph convolution with LSTM network to learn the relationships between roadways \citep{8917706}; (8) Attributed-augmented spatiotemporal graph convolutional network (\textbf{ASTGCN}), which integrate the external factors of dynamic and statistic factors into the spatiotemporal graph convolutional model \citep{9363197} and (9) Graph multi-head attention neural network (\textbf{GMHANN}), a latest prediction method which adopts encoder-decoder structure to capture the spatiotemporal dependencies based on AGRU and multi-head attention mechanism \citep{10234657}.   

\subsection{Settings of Hyperparameters and Computing Environment}
In our experiments, we configured the batch size to 33, the number of hidden units to 64, the learning rate to 0.01, the GCN layers to 2, the GRU layers to 3, and the dropout rate to 0.2. During the evaluation of prediction performance, we preserved the foundational hyper-parameter settings for both the GFEN and baseline methods. Furthermore, in the data processing phase, we utilized 80$\%$ of the historical observations for model training, while reserving 20$\%$ of the data for the test set.

The detailed settings of the computing environment are shown as follows. We conducted the experiments using a server with 16 CPU cores (Intel i7 13700k) and one GPU (RTX 4090). The version of Python is 3.7, we use Scikit-learn, Tensorflow, and Keras for the model construction.

\section{Results and Analysis} \label{sec-res}
\subsection{Prediction Performance}
\begin{figure}[h]
	\centering
	\begin{subfigure}[b]{0.38\textwidth}
		\includegraphics[width=\textwidth]{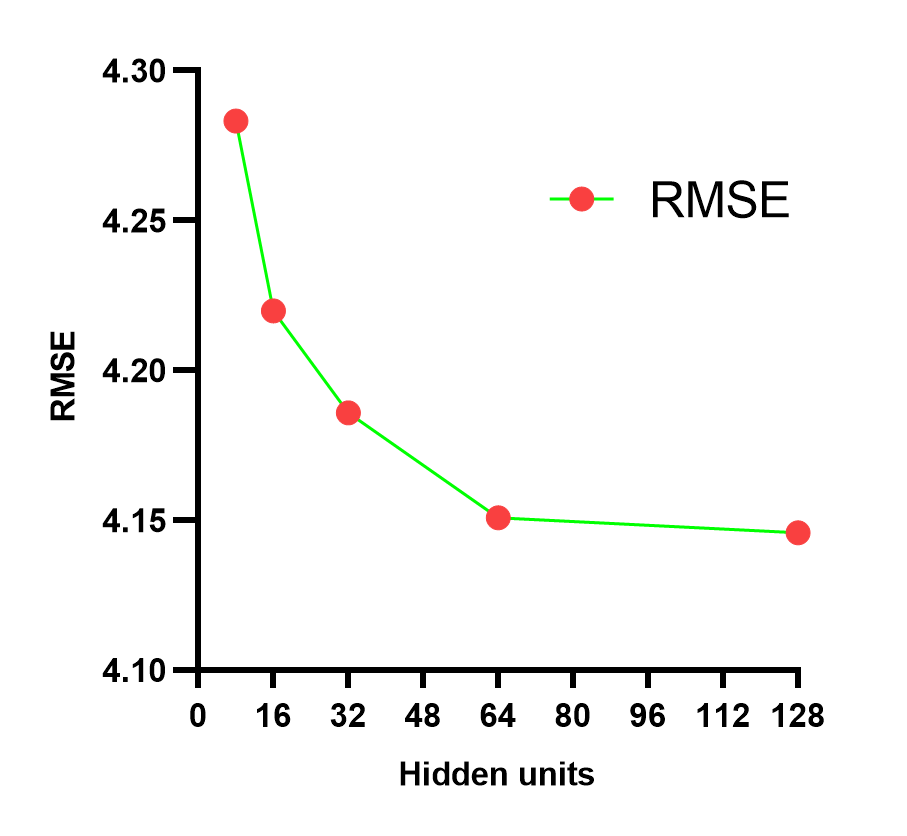}
		\label{fig:sub1}
	\end{subfigure}
	\begin{subfigure}[b]{0.38\textwidth}
		\includegraphics[width=\textwidth]{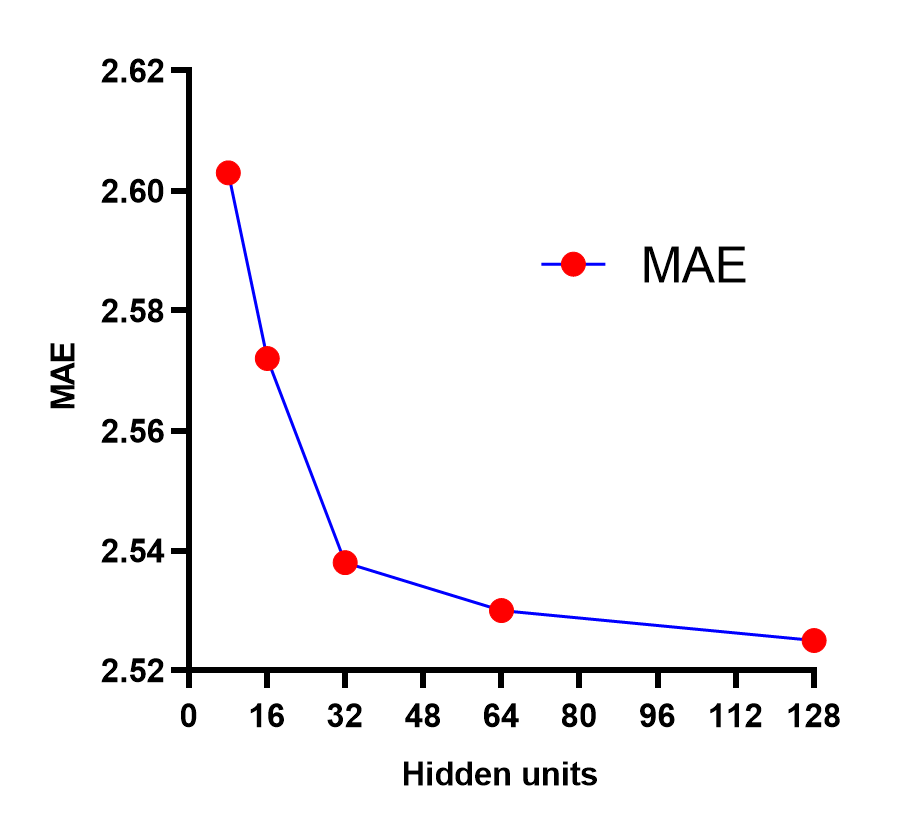}
		\label{fig:sub2}
	\end{subfigure}
	\caption{Prediction loss under different hidden units based on the Los-loop dataset.}
	\label{fig:9}
\end{figure}

\begin{figure}[h]
	\centering
	\begin{subfigure}[b]{0.38\textwidth}
		\includegraphics[width=\textwidth]{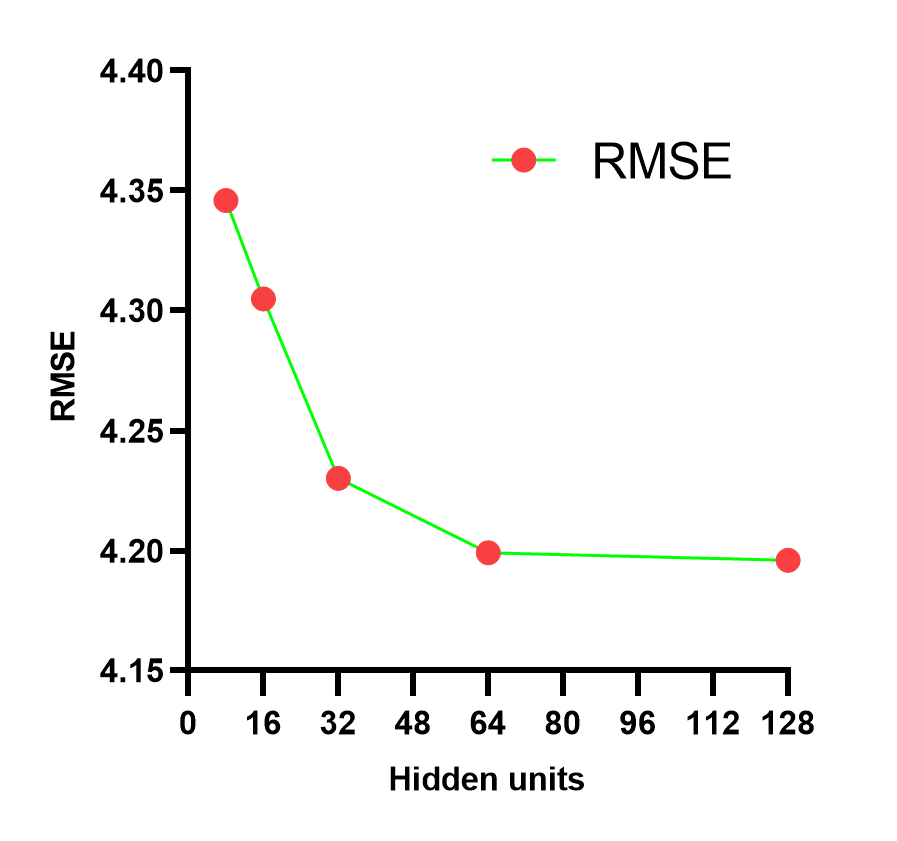}
		\label{fig:sub1}
	\end{subfigure}
	\begin{subfigure}[b]{0.38\textwidth}
		\includegraphics[width=\textwidth]{Losmae.png}
		\label{fig:sub2}
	\end{subfigure}
	\caption{Prediction loss under different hidden units based on the Seattle-loop dataset.}
	\label{fig:10}
\end{figure}

We first assess the prediction performance of the Graph Fusion Enhanced Network (GFEN) using two datasets. In our experiments, we evaluate the prediction performance of the model with varying numbers of hidden units. Figures \ref{fig:9} and \ref{fig:10} illustrate the changes in prediction accuracy across the two datasets. As the number of hidden units increases, prediction accuracy tends to improve, although the trend levels off when the number reaches 64. It is well-known that increasing the number of hidden units can lead to longer computational times. Therefore, considering both prediction accuracy and efficiency, we set the number of hidden units for GFEN to 64 when testing with the Los-loop dataset.


\begin{figure}[h]
	\centering
	\begin{subfigure}[b]{0.4\textwidth}
		\includegraphics[width=\textwidth]{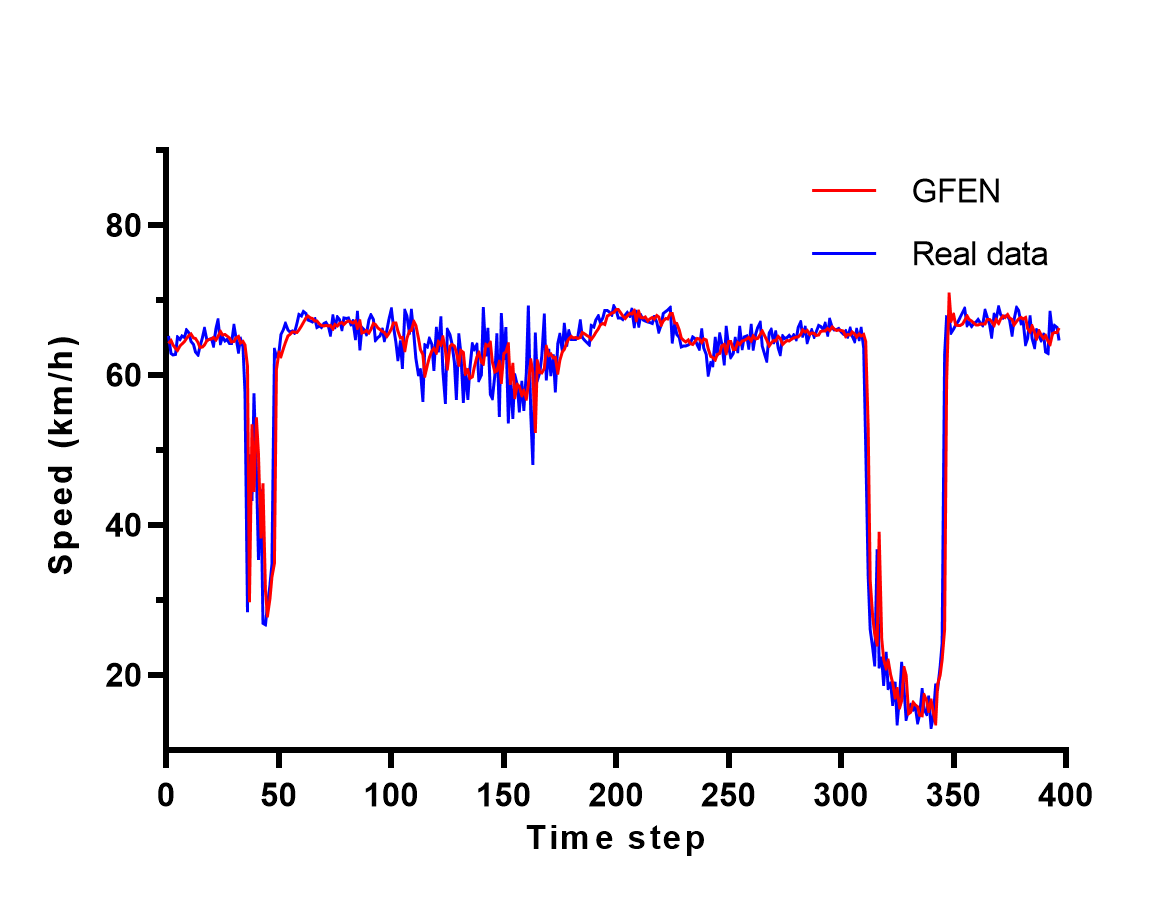}
            \caption{}
		\label{fig:sub1}
	\end{subfigure}
	\begin{subfigure}[b]{0.4\textwidth}
		\includegraphics[width=\textwidth]{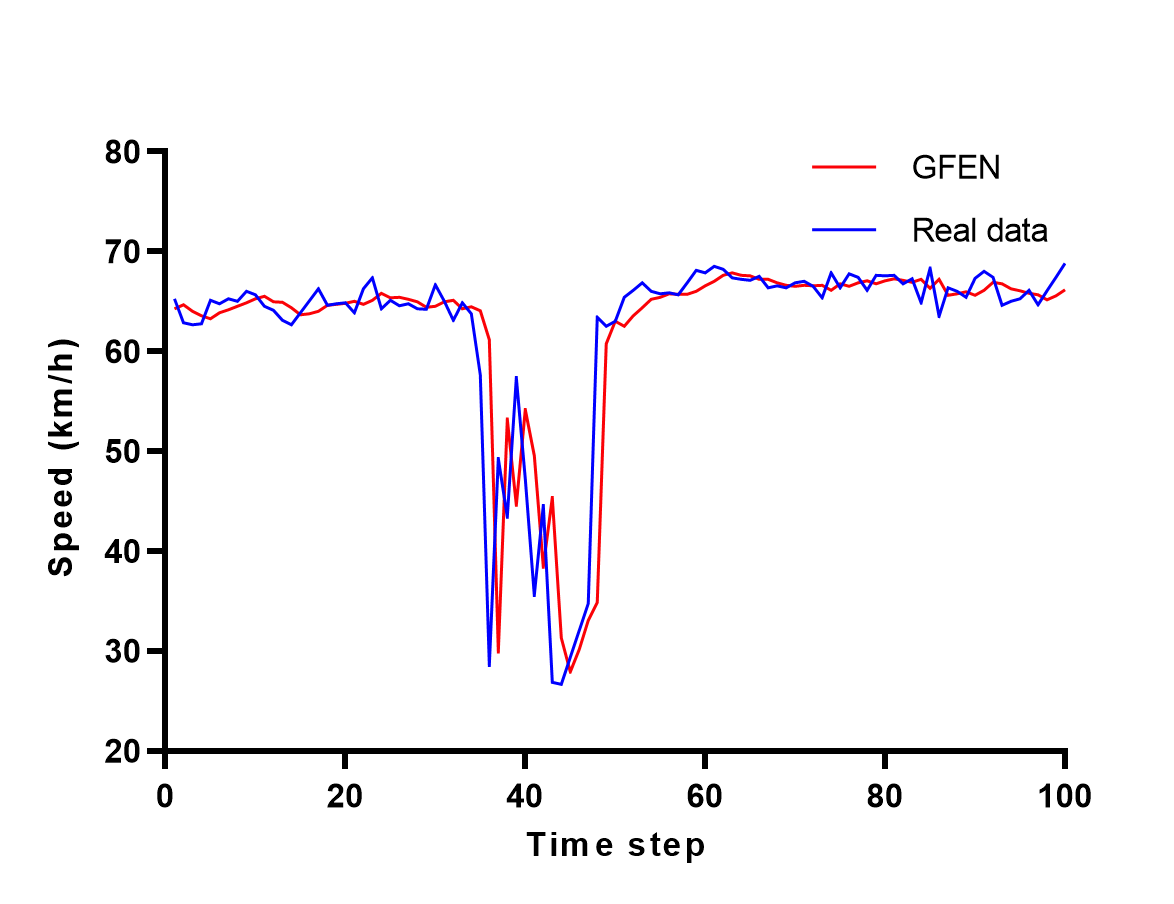}
            \caption{}
		\label{fig:sub2}
	\end{subfigure}
	\caption{Visualization results based on the Los-loop dataset with 64 hidden units. (a) Results over all time steps. (b) Part of the results.}
	\label{fig:11}
\end{figure}
\begin{figure}[h]
	\centering
	\begin{subfigure}[b]{0.4\textwidth}
		\includegraphics[width=\textwidth]{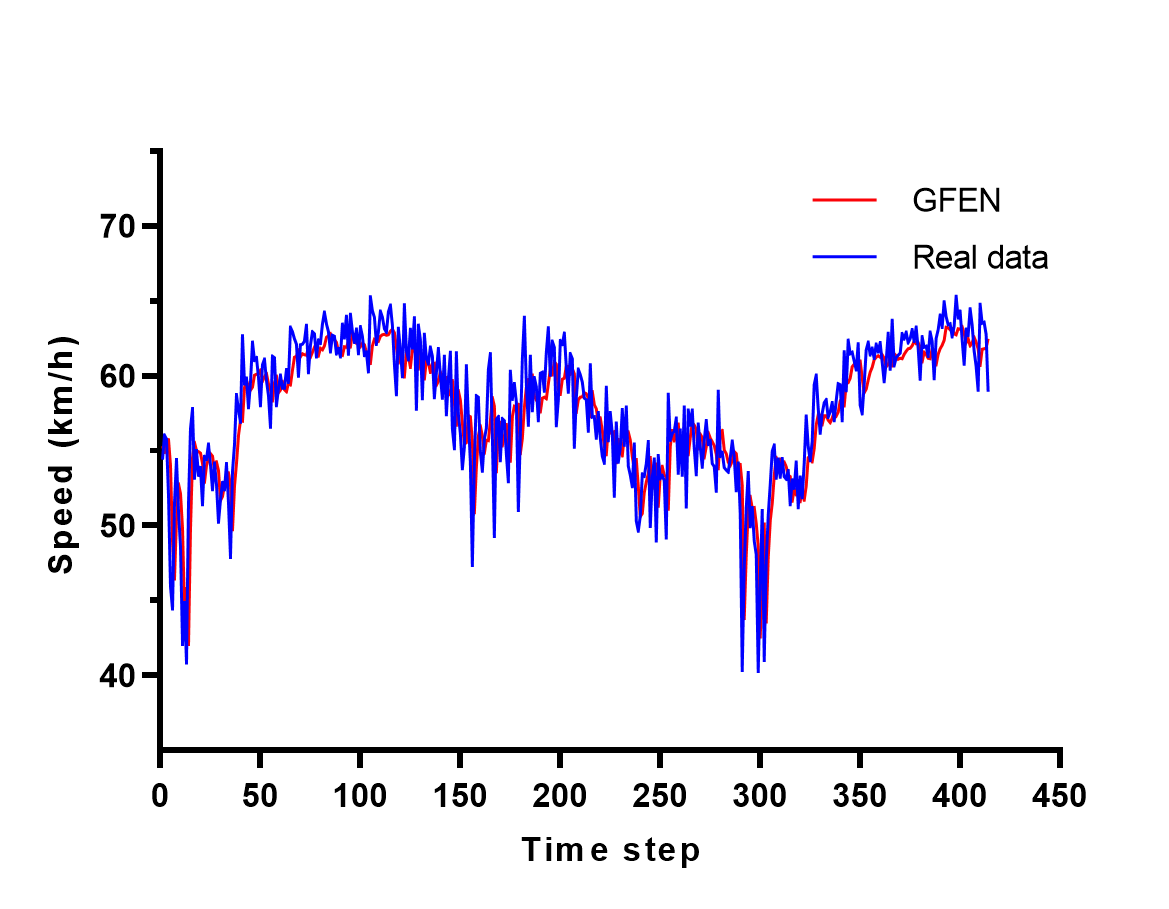}
            \caption{}
		\label{fig:sub1}
	\end{subfigure}
	\begin{subfigure}[b]{0.4\textwidth}
		\includegraphics[width=\textwidth]{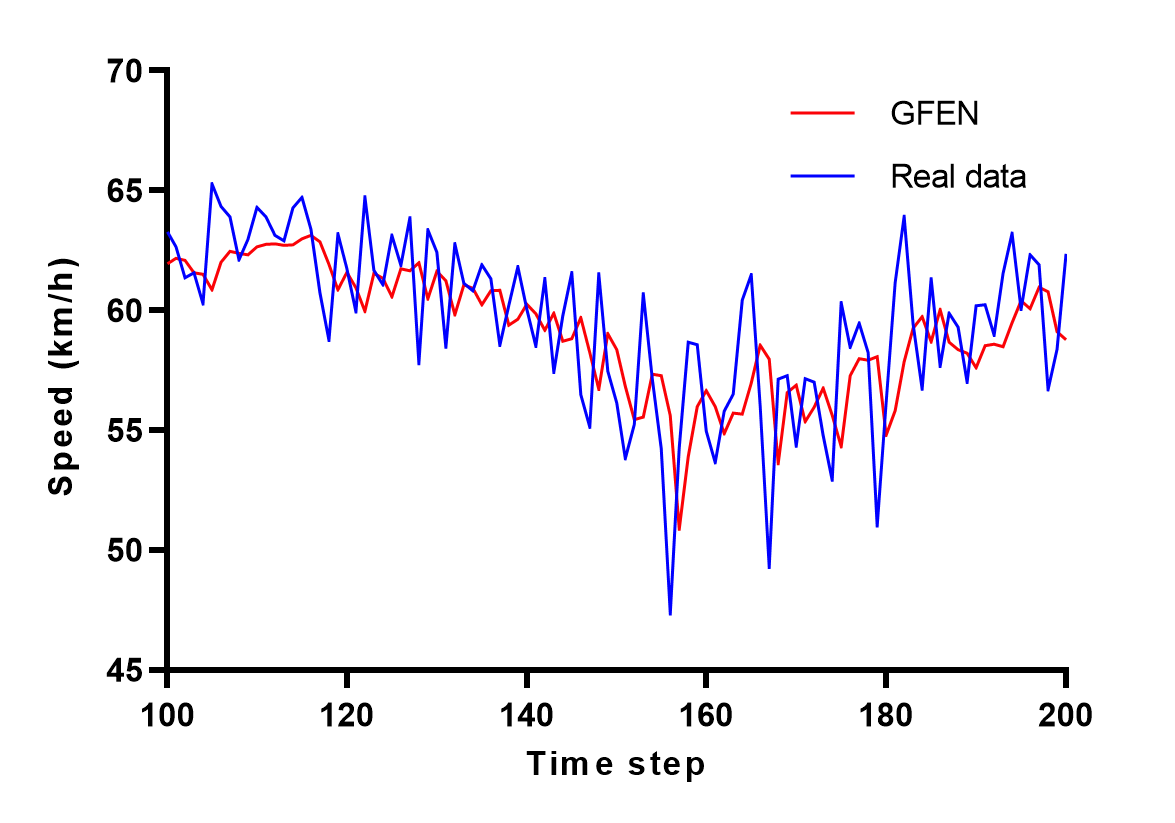}
            \caption{}
		\label{fig:sub2}
	\end{subfigure}
	\caption{Visualization results based on the Seattle-loop dataset with 64 hidden units. (a) Results over all time steps. (b) Part of the results.}
	\label{fig:12}
\end{figure}

Figures \ref{fig:11} and \ref{fig:12} present the prediction results of the Graph Fusion Enhanced Network (GFEN) based on the two datasets. Specifically, Figures \ref{fig:11}(a) and \ref{fig:12}(a) illustrate the prediction outcomes over the tested time steps. To facilitate a clearer comparison between the predicted results and the actual data, we have extracted the predictions for a subset of the time steps. As shown in Figures \ref{fig:11}(b) and \ref{fig:12}(b), the prediction results align closely with the ground truth, demonstrating the effectiveness of our GFEN model. Moreover, when comparing the predicted results to the ground truth data, we observe that the prediction curves obtained by GFEN are smoother for both datasets. However, the model's performance diminishes when applied to local traffic data exhibiting significant fluctuations. This preliminary analysis suggests that the GCN-based models may smooth out these fluctuations due to their design, which utilizes a constant moving filter for modeling spatial features, thereby dampening the peaks in the predictions.

Additionally, we have incorporated the Topological Spatiotemporal Graph Fusion (TSTGF) technique and the Enhanced Data Correlation (EDC) technique into the hybrid model comprising GCN and GRU. This integration enables us to separately evaluate the effectiveness of each technique by analyzing their impacts on prediction accuracy. Finally, we assess the GFEN network, which combines both techniques, to evaluate the overall improvements achieved by the model introduced in this paper. As illustrated in Table \ref{tab:1}, both techniques contribute to enhanced prediction accuracy compared to the baseline model. The advantages become even more pronounced when utilizing the GFEN network, further emphasizing the effectiveness of both techniques.

\begin{table*}[t]
	\centering
	\setlength{\tabcolsep}{5pt}{
		\renewcommand\arraystretch{1.5}{
			\caption{Prediction performance of applying different proposed techniques}
                \label{tab:1}
			\begin{tabular}{c c c c c}\hline
				Dataset & \multicolumn{2}{c}{Los-loop dataset} & \multicolumn{2}{c}{Seattle-loop dataset}\\ \hline
				Method & RMSE & MAE & RMSE & MAE\\ \hline
				GCN+GRU & 5.0200 & 3.3667 & 4.6907 & 3.2031 \\ 
				(GCN+GRU)+TSTGF & 4.3026 & 2.6401 & 4.3775 & 2.8669 \\ 
				(GCN+GRU)+EDC & 4.6060 & 2.9555 & 4.4334 & 2.9394 \\
				GFEN & 4.1517 & 2.5303 & 4.1996 & 2.7441 \\ \hline 
			\end{tabular}
		}
	}
\end{table*}

\begin{table}[h]
	\centering
	\setlength{\tabcolsep}{1.8mm}{
		\renewcommand\arraystretch{1.9}{
			\caption{Prediction performance based on the Los-loop dataset.}
                \label{tab:2}
			\begin{tabular}{c c c c c c c c c c c}\hline
				Method & GFEN & GMHANN & ASTGCN & TGC-LSTM & A3T-GCN & T-GCN & LSTM & SVR & HA & ARIMA\\ \hline
				$RMSE$ & \textbf{4.1517} & 4.6597 & 4.5536 & 4.8700 & 4.8360 & 5.0200 & 9.6028 & 11.1333 & 7.3067 & 10.0625\\ 
				$MAE$ & \textbf{2.5303} & 3.0987 & 2.8653 & 3.1656 & 3.2037 & 3.3667 & 6.7561 & 7.1441 & 3.8782 & 7.7306\\ 
				$ACC$ & \textbf{0.9294} & 0.9212 & 0.9256 & 0.9171 & 0.9077 & 0.9146 & 0.8026 & 0.8106 & 0.8756 & 0.8272\\ 
				$R^2$ & \textbf{0.9095} & 0.8993 & 0.9093 & 0.8755 & 0.9077 & 0.8677 & 0.4935 & 0.3495 & 0.7225 & *\\ 
				$VAR$ & \textbf{0.9095} & 0.8758 & 0.8892 & 0.8756 & 0.8773 & 0.8702 & 0.5078 & 0.3514 & 0.7225 & 0.0015\\ \hline 
			\end{tabular}
		}
	}
\end{table}
\begin{table}[h]
	\centering
	\setlength{\tabcolsep}{1.8mm}{
		\renewcommand\arraystretch{1.9}{
			\caption{Prediction performance based on the Seattle-loop dataset.}
                \label{tab:3}
			\begin{tabular}{c c c c c c c c c c c}\hline
				Method & GFEN & GMHANN & ASTGCN & TGC-LSTM & A3T-GCN & T-GCN & LSTM & SVR & HA & ARIMA\\ \hline
				$RMSE$ & \textbf{4.1996} & 4.2736 & 4.3166 & 4.3812 & 5.0562 & 4.6907 & 8.3794 & 10.7602 & 7.3067 & 10.8712\\ 
				$MAE$ & \textbf{2.7441} & 2.8516 & 2.9076 & 2.9530 & 3.4987 & 3.2031 & 5.3749 & 7.0126 & 6.8031 & 8.5355\\ 
				$ACC$ & \textbf{0.9272} & 0.9262 & 0.9140 & 0.9240 & 0.9123 & 0.9186 & 0.8373 & 0.8134 & 0.8821 & 0.8097\\ 
				$R^2$ & \textbf{0.8839} & 0.8765 & 0.8723 & 0.8736 & 0.8317 & 0.8551 & 0.5208 & 0.2377 & 0.6966 & *\\ 
				$VAR$ & \textbf{0.8841} & 0.8712 & 0.8717 & 0.8736 & 0.8317 & 0.8562 & 0.5210 & 0.2532 & 0.6967 & *\\ \hline 
			\end{tabular}
		}
	}
\end{table}

\subsection{Prediction Performance Comparison}
To assess the superiority of the Graph Fusion Enhanced Network (GFEN), we compare its prediction accuracy against several baseline methods. Specifically, to ensure the reliability of the experimental results, we maintain the same number of hidden units for GFEN and baseline methods.

First, we evaluate all methods using the Los-loop dataset. Table \ref{tab:2} presents the prediction accuracy of the GFEN model alongside various baseline methods, with each result displayed to four decimal places and an asterisk $*$ indicating values below zero. Our findings indicate that the GFEN model achieves the highest prediction performance, demonstrating its effectiveness for traffic prediction tasks. For instance, the Mean Squared Error (MSE) and Mean Absolute Error (MAE) of the GFEN model show improvements of 10.9$\%$ and 18.3$\%$, respectively, compared to the GMHANN model. Furthermore, when compared to the ASTGCN model, GFEN exhibits enhancements in RMSE and MAE of approximately 8.5$\%$ and 11.6$\%$, respectively. Additionally, the RMSE and MAE of GFEN are reduced by 14.7$\%$ and 20.1$\%$ compared to the TGC-LSTM model. Overall, GFEN demonstrates a notably superior prediction accuracy compared to other baseline methods, highlighting its robust performance.

\begin{figure}[h]
	\centering
	\begin{subfigure}[b]{0.45\textwidth}
		\includegraphics[width=\textwidth]{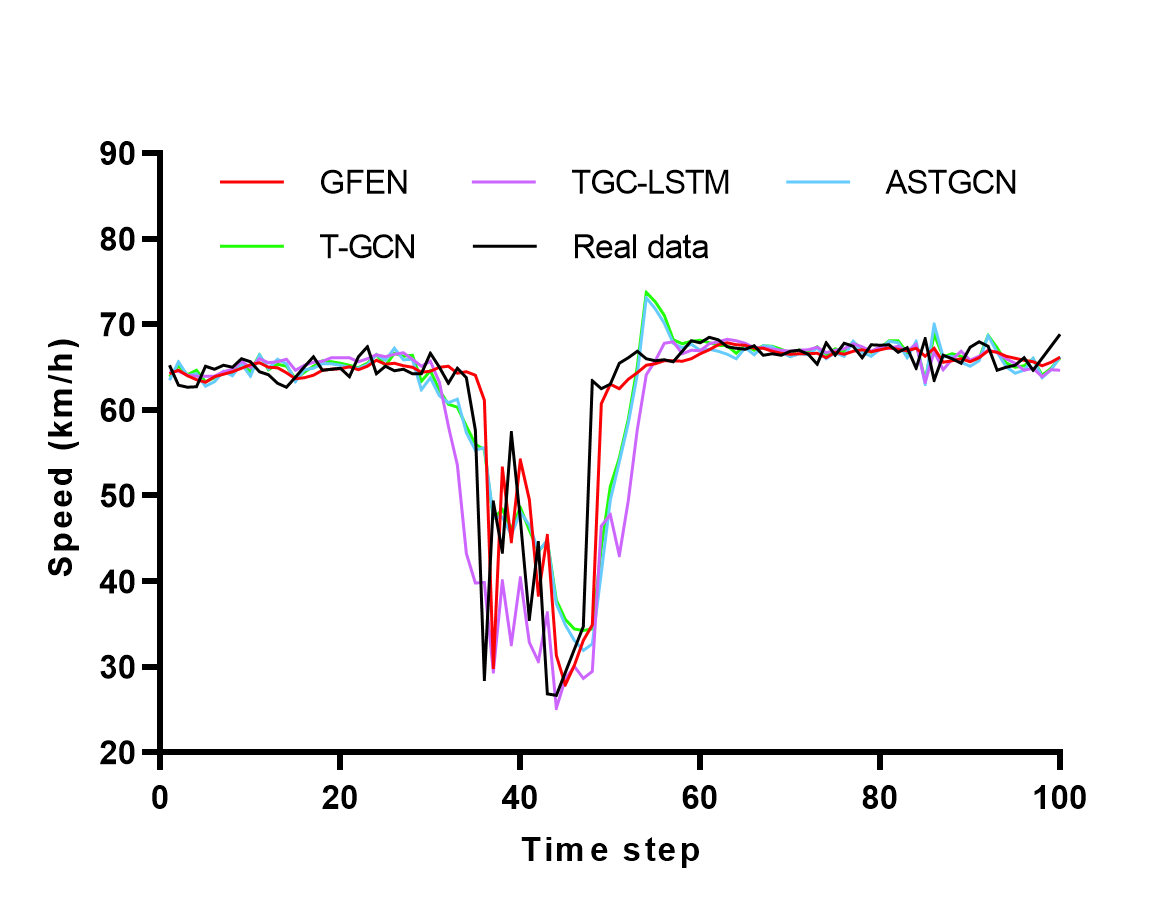}
            \caption{}
		\label{fig:sub1}
	\end{subfigure}
	\begin{subfigure}[b]{0.45\textwidth}
		\includegraphics[width=\textwidth]{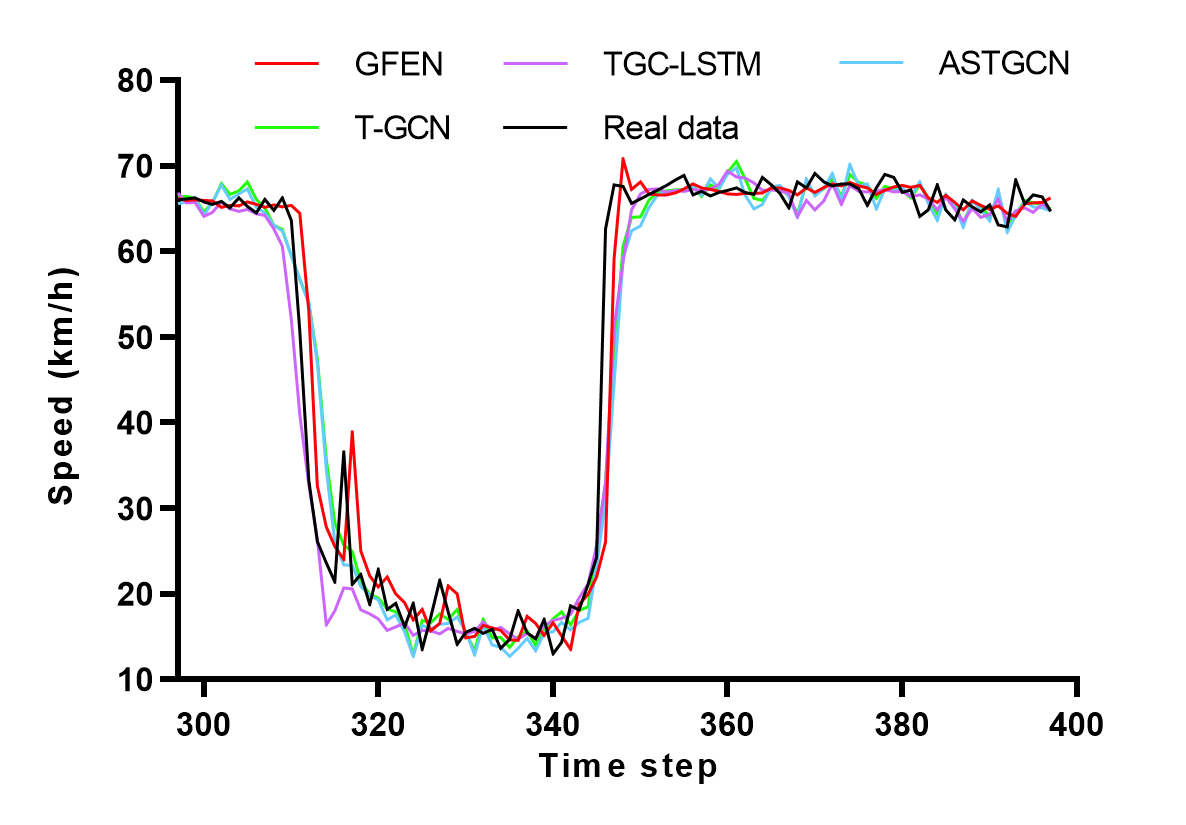}
            \caption{}
		\label{fig:sub2}
	\end{subfigure}
	\caption{Prediction results of GFEN and baseline methods with Los-loop dataset. (a) Results from time step 1 to 100. (b) Results from time step 300 to 400.}
	\label{fig:13}
\end{figure}
\begin{figure}[h]
	\centering
	\begin{subfigure}[b]{0.45\textwidth}
		\includegraphics[width=\textwidth]{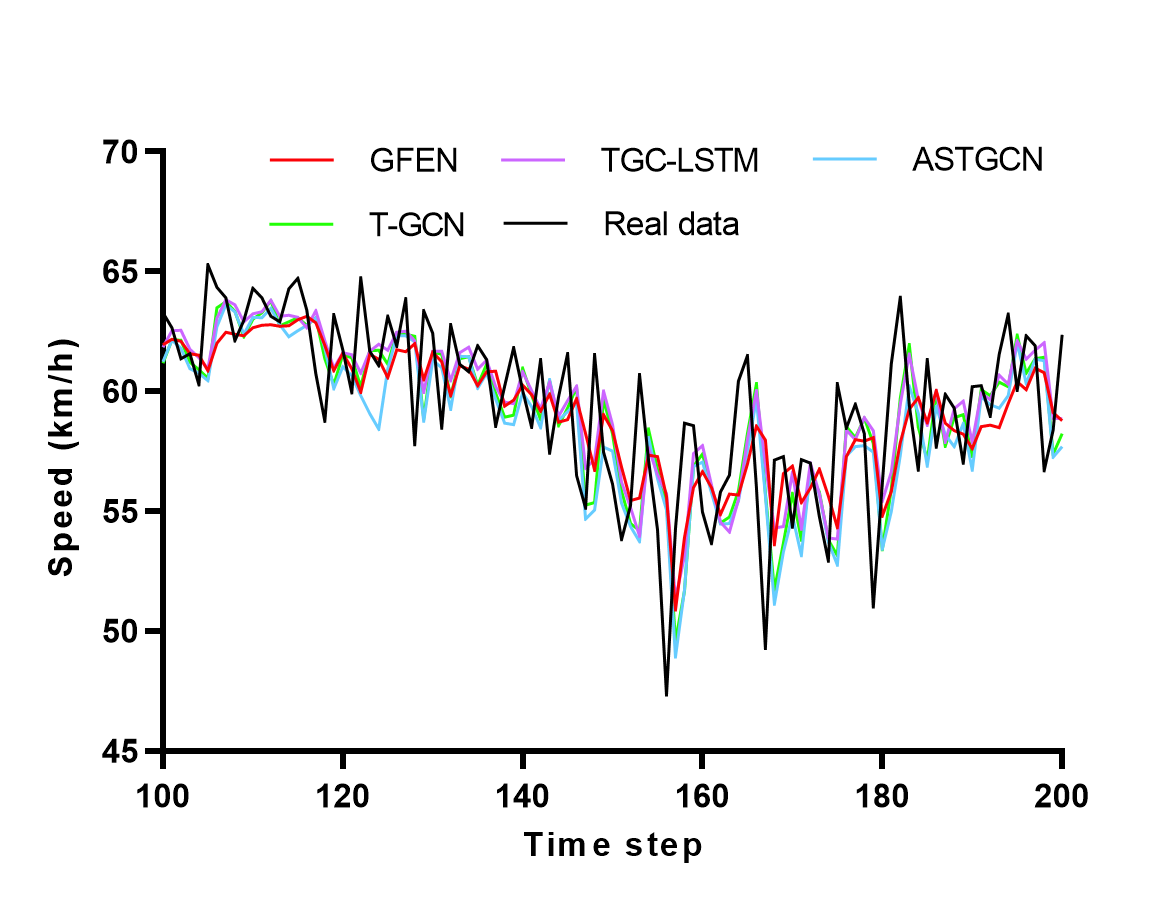}
            \caption{}
		\label{fig:sub1}
	\end{subfigure}
	\begin{subfigure}[b]{0.45\textwidth}
		\includegraphics[width=\textwidth]{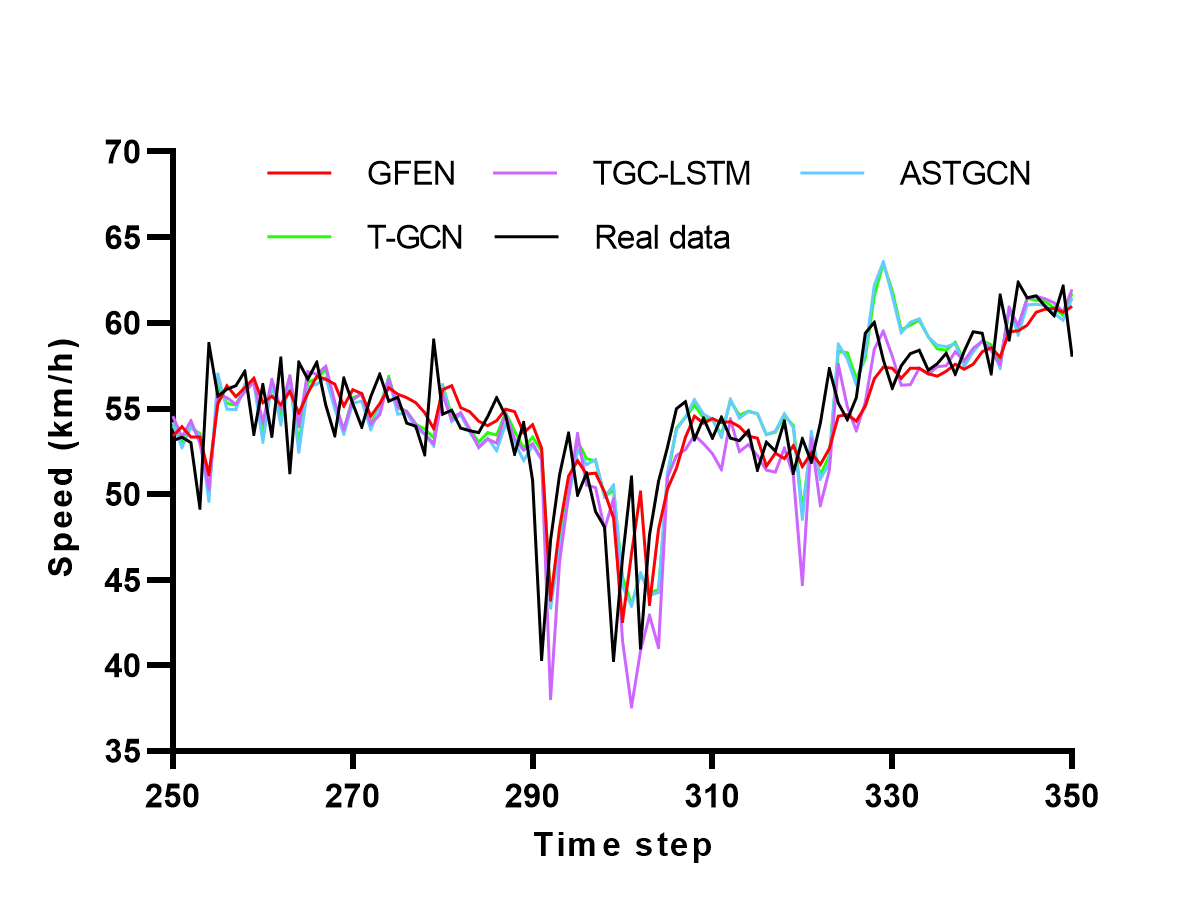}
            \caption{}
		\label{fig:sub2}
	\end{subfigure}
	\caption{Prediction results of GFEN and baseline methods with Los-loop dataset. (a) Results from time step 1 to 100. (b) Results from time step 300 to 400.}
	\label{fig:14}
\end{figure}
Second, we evaluate all methods using the Seattle-loop dataset. Table \ref{tab:3} presents the prediction performance based on this dataset. From Table \ref{tab:3}, it is evident that the GFEN model retains significant advantages over recent hybrid prediction models. Specifically, compared to the GMHANN model, GFEN demonstrates improvements in Root Mean Squared Error (RMSE) and Mean Absolute Error (MAE) of 1.7$\%$ and 1.9$\%$, respectively. Additionally, GFEN outperforms the ASTGCN model with improvements of 2.7$\%$ in RMSE and 5.6$\%$ in MAE. Furthermore, when compared to the TGC-LSTM model, GFEN achieves reductions in RMSE and MAE of 4.1$\%$ and 7.0$\%$, respectively.

From Table \ref{tab:2} and Table \ref{tab:3}, it can be seen that GFEN shows superiority over the baseline methods based on the two datasets. In addition, we find that among all traffic prediction models, hybrid deep learning-based models such as GFEN and GMHANN, which capture the spatiotemporal dependencies during model training, generally have better prediction performance than other baselines. This is mainly because single time series prediction models cannot model the spatial features of traffic data and ignore the spatial correlation among traffic conditions at multi-located road sections.

Furthermore, to evaluate the effectiveness of GFEN in processing anomalous traffic data, we extract the data with large fluctuations and test the prediction performance of GFEN. Specifically, we select several hybrid deep learning-based models that perform well in Table \ref{tab:2} and Table \ref{tab:3}, including TGC-LSTM, ASTGCN, and T-GCN as the baselines in this stage. From both two datasets, we select two sections of data with large fluctuations and compare the prediction results of the tested methods, as shown in Figure \ref{fig:13} and Figure \ref{fig:14} respectively. The two figures indicate that the prediction results of GFEN show a stronger similarity with the real data if there are fluctuations in some data. Also, all methods predict poorly at local minima/maxima. However, the error of prediction results obtained by GFEN is the smallest among all methods in most cases.

In summary, GFEN shows superiority over the baseline methods under most of the cases. With consideration of the data containing large fluctuations and anomalous data points, the prediction results of GFEN are in better line with the ground truth data, which indicates the advantage of our model in predicting non-stationary and anomalous traffic data. 
\subsection{Computational Time and Convergence Speed Comparison}
Time complexity is an important factor in ensuring the model efficiency. In experiments, we evaluate the time complexity of GFEN in terms of computational time and model convergence speed.

First, we test the total computational time of GFEN during model training under different numbers of hidden units. As shown in Table \ref{tab:4}, for both two real-world datasets, we find that the computational time will increase with the increasing number of hidden units.

Furthermore, we compare the computation time of GFEN with baseline methods, it should be noted that all the settings of hyper-parameters remain consistence during the comparison stage. As shown in Table \ref{tab:5}, the computational time of hybrid prediction models including GFEN, GMHANN, and TGC-LSTM. A3T-GCN and T-GCN are much longer than single time series prediction models such as LSTM, HA, and ARIMA. This is mainly because the hybrid models employ complex structures to model the spatial and temporal features, which increases the computational time for higher accuracy. From the stage of results comparison, we find that the prediction accuracy of hybrid models is much higher than traditional time series prediction models, indicating the worth of additional computational time of hybrid prediction models.

Specifically, the computational time of the hybrid prediction models is not much different, where the computational time of A3T-GCN is the highest and GMHANN is the lowest. Furthermore, we test the convergence speed and the computational time for each training epoch of the hybrid prediction models. Here, we provide the definition of the model convergence. In each training epoch, we output the training error given by RMSE which retains three decimal places. If the training error first remains the same in two consecutive training epochs, then we consider the model converges in this training epoch. Table \ref{tab:6} shows the convergence training epoch for each method and the computational time of each epoch, where $>300$ denotes the model cannot reach convergence within 300 training epochs. 
\begin{table}[h]
	\centering
	\setlength{\tabcolsep}{10mm}{
		\renewcommand\arraystretch{1.5}{
			\caption{Total computational time of GFEN under different numbers of hidden units.}
                \label{tab:4}
			\begin{tabular}{c c c}\hline
				Hidden units & Los-loop dataset/s & Seattle-loop dataset/s\\ \hline
				8 & 346 & 430 \\ 
				16 & 731 & 1139 \\ 
				32 & 1035 & 1867 \\ 
				64 & 2324 & 3560 \\ 
				128 & 5021 & 6762 \\ \hline 
			\end{tabular}
		}
	}
\end{table}

\begin{table}[h]
	\centering
	\setlength{\tabcolsep}{16mm}{
		\renewcommand\arraystretch{1.5}{
			\caption{Computational time of tested methods based on the two datasets.}
                \label{tab:5}
			\begin{tabular}{c c c}\hline
				Methods & Los-loop dataset/s & Seattle-loop dataset/s\\ \hline
				GFEN & 2324 & 3560 \\ 
				GMHANN & 2157 & 3072 \\
				ASTGCN & 2258 & 3262 \\
				TGC-LSTM & 2249 & 3116 \\ 
				A3T-GCN & 2671 & 3784 \\ 
				T-GCN & 3131 & 3270 \\ 
				LSTM & 1851 & 2767 \\ 
				SVR & 17 & 24 \\ 
				HA & 0.3 & 1.0 \\ 
				ARIMA & 52 & 86 \\ \hline 
			\end{tabular}
		}
	}
\end{table}

\begin{table*}[t]
	\centering
	\setlength{\tabcolsep}{5pt}{
		\renewcommand\arraystretch{1.5}{
			\caption{Convergence epoch and training time for hybrid models.} 
                \label{tab:6}
			\begin{tabular}{c c c c c}\hline
				Dataset & \multicolumn{2}{c}{Los-loop dataset} & \multicolumn{2}{c}{Seattle-loop dataset}\\ \hline
				Method & Convergence epoch & Training time (s/epoch)& Convergence epoch & Training time (s/epoch)\\ \hline
				GFEN & 113 & 7.7 & 62 & 11.8 \\ 
				GMHANN & 231 & 7.2 & 119 & 10.2 \\ 
				ASTGCN & 254 & 7.5 & 162 & 10.8 \\
				TGC-LSTM & 267 & 7.4 & 193 & 10.3 \\ 
				A3T-GCN & $>300$ & 8.9 & 206 & 10.9 \\ 
				T-GCN & $>300$ & 10.4 & $>300$ & 1867 \\ \hline 
			\end{tabular}
		}
	}
\end{table*}

From Table \ref{tab:6}, we find that GFEN reaches convergence much faster than the other hybrid prediction models. Among all methods, A3T-GCN cannot reach convergence when tested with the Los-loop dataset and T-GCN cannot converge based on both datasets within 300 training epochs. Specifically, compared with the GMHANN model which reaches convergence within the lowest epochs among all baseline methods, the number of training epochs required for GFEN to reach convergence is half or less than that of GMHANN. When the training time for each epoch is similar, GFEN needs fewer training epochs to reach convergence compared with other baseline methods, which represents that the computational time of GFEN is far less than the other hybrid prediction models, indicating the high efficiency of the GFEN model in traffic prediction tasks. 

\subsection{Robustness}
Noise is inevitable during the data-collecting process. Hence, we test the robustness of the GFEN model, which aims to test the impact of noise on our model.

\begin{figure}[h]
	\centering
	\subfloat[Adding Gaussian noise on the Los-loop dataset]{
		\includegraphics[scale=0.6]{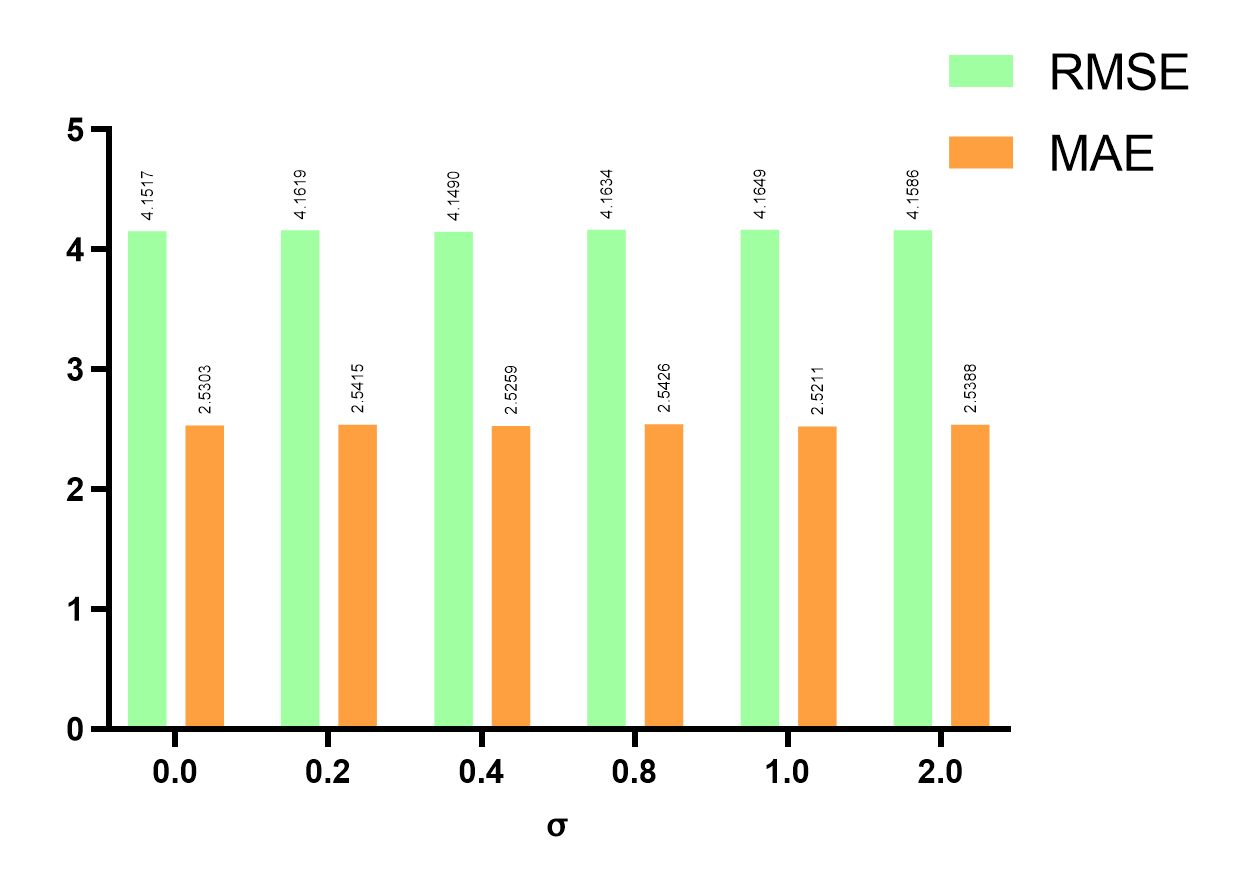}}
	\subfloat[Adding Gaussian noise on the Seattle-loop dataset]{
		\includegraphics[scale=0.6]{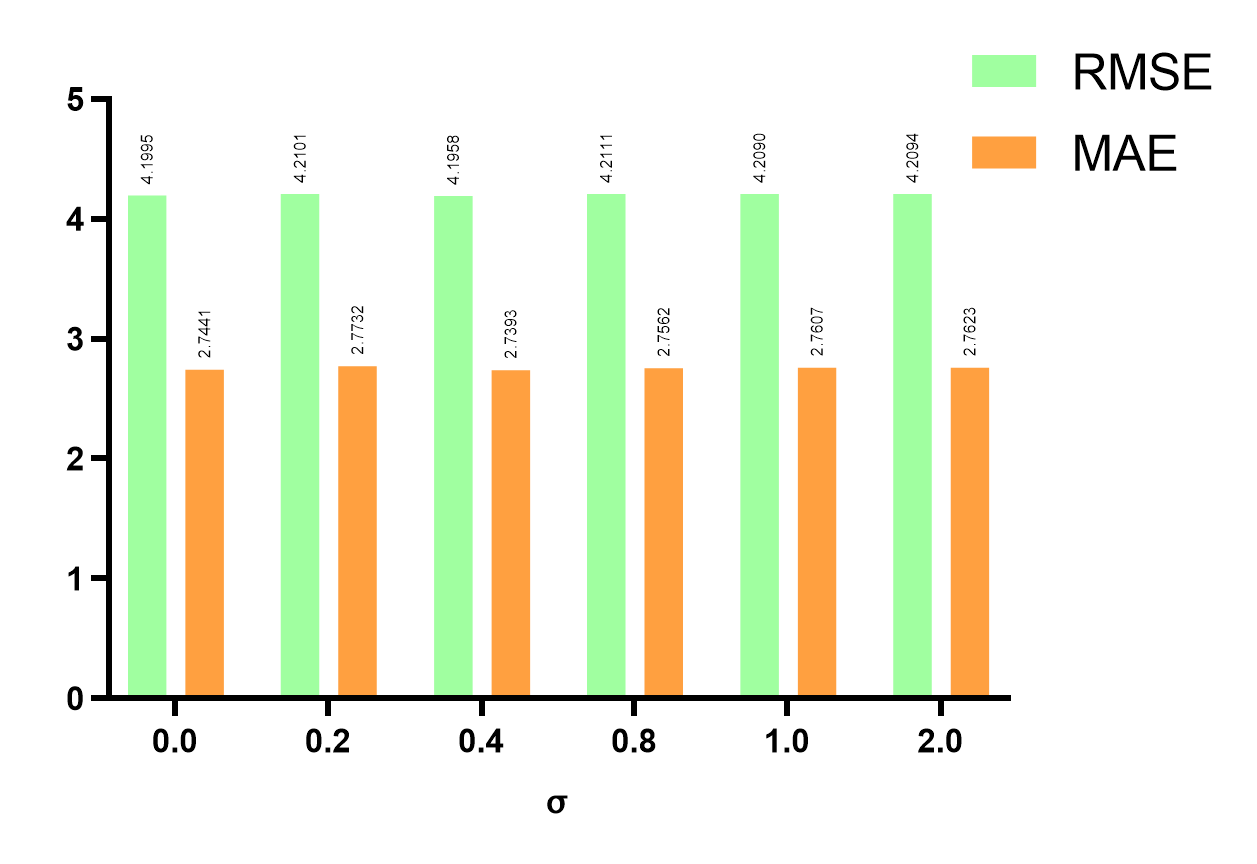}}
	\\
	\subfloat[Adding Poisson noise on the Los-loop dataset]{
		\includegraphics[scale=0.6]{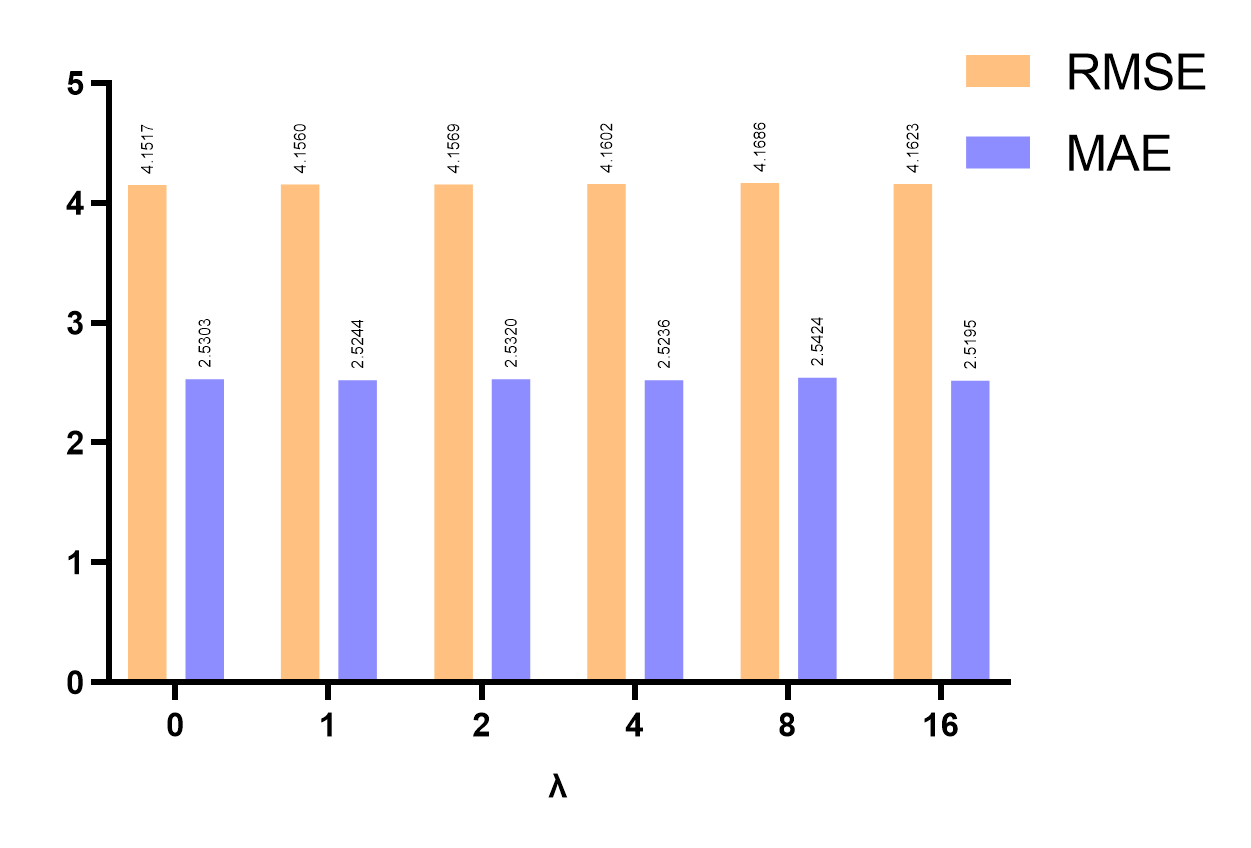}}
	\subfloat[Adding Poisson noise on the Seattle-loop dataset]{
		\includegraphics[scale=0.6]{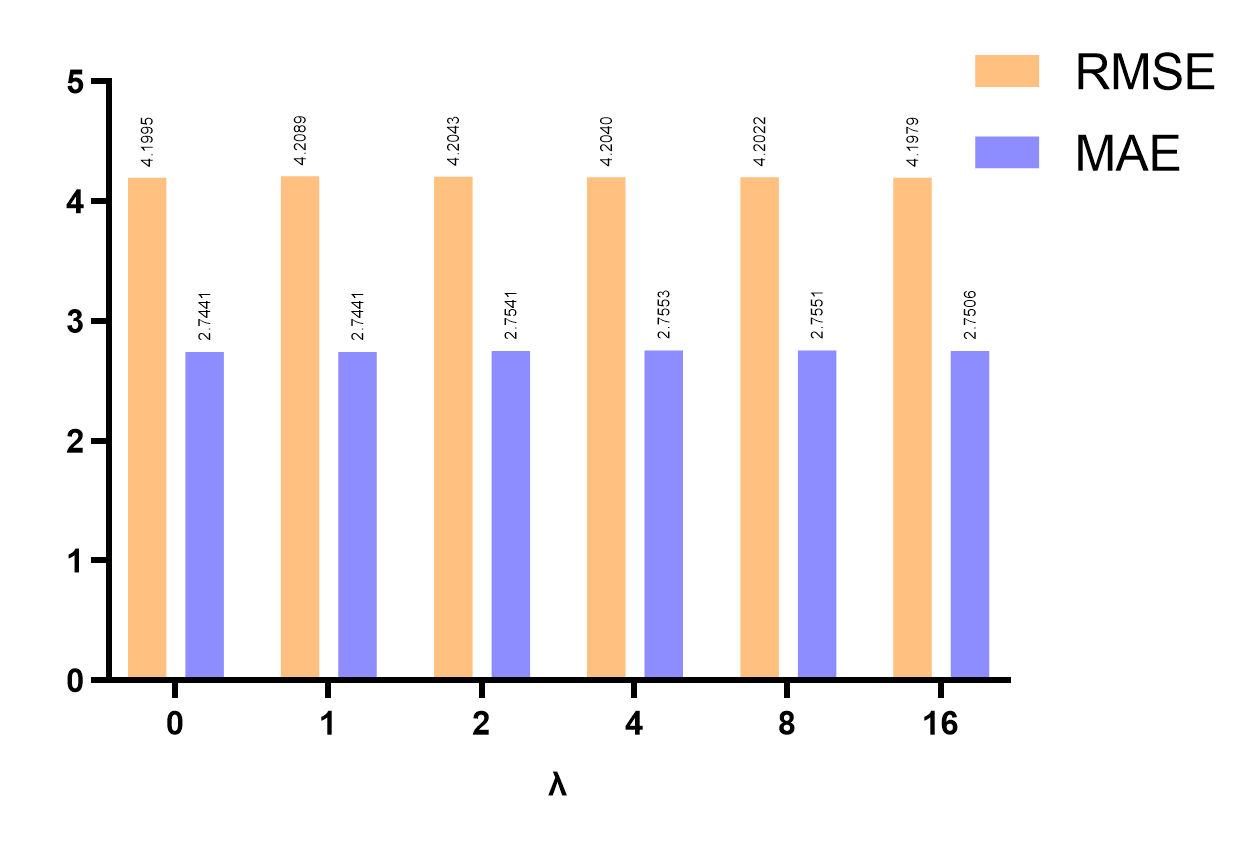}}
	\caption{Robustness test.}
	\label{fig:15}
\end{figure}

In our experiments, two common noises are added to the historical observations. The first type of noise obeys the Gaussian distribution $N\in(0, \sigma^2)$, where $\sigma\in(0.2, 0.4, 0.8, 1, 2)$. The second obeys the Poisson distribution $P(\lambda)$, where $\lambda\in(1, 2, 4, 8, 16)$. Then, the value of the noise matrices is normalized to be between 0 and 1. As shown in Figure \ref{fig:15}, where the horizontal denotes the parameters of the noise model and the vertical axis represents the change of prediction accuracy given by the two evaluation metrics, we find that the change of loss is small with different distributions of noise. Hence, GFEN is robust and can handle the high noise problem.

\section{Conclusion} \label{sec-con}
In this paper, we propose a Graph Fusion Enhanced Network (GFEN) for predicting network-level traffic speed, particularly when historical observations contain anomalous data. GFEN is implemented in two main steps: (a) A hybrid mathematical-transformer model combined with \emph{k}-th order difference operation and Enhanced Data Correlation (EDC) approach are utilized to convert randomly historical observations into stationary time series, thereby minimizing the negative impact of non-stationarity and data anomalism during training. (b) A multi-graph fusion technique is introduced, which integrates the spatiotemporal correlations extracted from historical observations with the topological network structure, allowing for a deep capture of spatiotemporal dependencies. Through our experiments, we demonstrate that GFEN achieves superior prediction accuracy and lower computational time compared to state-of-the-art methods. Moreover, GFEN shows strong robustness in addressing high noise issues.

Given the superior effectiveness and efficiency of GFEN, future research could explore its application in other fields involving spatiotemporal data prediction tasks, such as vehicle trajectory prediction, where only minor obstacles are anticipated. Furthermore, we can develop more effective operations to address problems arising from the non-stationarity of time series, with the goal of adaptively enhancing the accuracy of parameter estimation
\section{Acknowledgment} \label{sec-ack}
This work is supported in part by the National Natural Science Foundation of China (Grant No. 62372384) and the Research Development Fund of XJTLU under Grant RDF-21-02-082.

\bibliographystyle{unsrt}
\bibliography{Bibliography}

\end{document}